\documentclass[lettersize,journal]{IEEEtran}
\usepackage{amsmath,amsfonts}
\usepackage{algorithm}
\usepackage{algpseudocode}
\usepackage{array}
\usepackage[caption=false,font=normalsize,labelfont=sf,textfont=sf]{subfig}
\usepackage{textcomp}
\usepackage{bm}
\usepackage{stfloats}
\usepackage{url}
\usepackage{verbatim}
\usepackage{graphicx,amssymb}
\usepackage{cite}
\usepackage{booktabs}   
\usepackage{multirow}   
\usepackage{tabularx} 
\usepackage{adjustbox}
\usepackage[colorlinks,
            linkcolor=blue,
            anchorcolor=blue,
            citecolor=blue
            ]{hyperref}
\usepackage{xcolor}
\usepackage{booktabs}
\usepackage{nameref,flushend}
\usepackage{orcidlink}
\begin{document}

\title{A Physics-Informed Hierarchical Neural Network for\\ Microwave Scattering Analysis of 3D PEC Targets}

\author{
Rui Zhu\,\orcidlink{0009-0001-1035-0886},~\IEEEmembership{Student Member,~IEEE},
Yuexing Peng\,\orcidlink{0000-0002-6237-8995},~\IEEEmembership{Senior Member,~IEEE},\\
George C. Alexandropoulos\,\orcidlink{0000-0002-6587-1371},~\IEEEmembership{Senior Member,~IEEE},
and Wenbo Wang\,\orcidlink{0000-0002-0911-3189},~\IEEEmembership{Senior Member,~IEEE}
\thanks{This work was supported in part by the National Natural Science Foundation of China under Grant 62371066 and in part by the Science and Technology Innovation Program of Xiongan New Area under Grant 2025XAGG0052. \emph{(Corresponding author: Yuexing Peng)}}
\thanks{R. Zhu, Y. Peng, and W. Wang are with the Key Laboratory of Universal Wireless Communication, Ministry of Education, School of Information and Communication Engineering, Beijing University of Posts and Telecommunications, Beijing 100876, China (e-mail: \{rayyyy, yxpeng, wbwang\}@bupt.edu.cn).}
\thanks{G. C. Alexandropoulos is with the Department of Informatics and Telecommunications, National and Kapodistrian University of Athens, Athens 16122, Greece (e-mail: alexandg@di.uoa.gr).}
}

\maketitle

\begin{abstract}
Accurate modeling of scattering from three-dimensional (3D) perfectly electrically conducting (PEC) targets at microwave frequencies constitutes a fundamental objective in computational electromagnetics, particularly for radar cross section (RCS) prediction and microwave scattering analysis. Classical solvers, such as the method of moments and the Multilevel Fast Multipole Algorithm (MLFMA), although provide high physical fidelity, they become costly under scenarios of repeated queries involving many incidence
configurations or frequencies, whereas purely data-driven
surrogates often lack accuracy on geometrically complex
targets. This paper proposes a U-shaped physics-informed artificial neural network
(U-PINet) for 3D microwave scattering analysis.
Inspired by the near-far field decomposition of MLFMA, U-PINet
combines a near-field graph encoder, parameterized by
learnable univariate basis functions, with a hierarchical
multi-scale fusion module organized on an octree partition.
The proposed network is trained against a discretized residual of the electric-field integral equation at surface
collocation points, without requiring reference current
labels. Experiments on canonical and geometrically complex 3D PEC targets, conducted under multiple frequency and polarization configurations and assessed through bistatic RCS reconstruction, showcase that U-PINet outperforms representative physics-informed baselines, and yields substantial runtime savings over the classical MLFMA solver under repeated-query scenarios.
\end{abstract}


\section{Introduction}
\IEEEPARstart{E}{lectromagnetic} (EM) scattering modeling
of three-dimensional (3D) targets at microwave and
millimeter-wave frequencies is a canonical problem in
computational EM (CEM), underpinning a
broad range of microwave engineering tasks such as radar
cross section (RCS)
prediction~\cite{balanis2012advanced}, microwave imaging
and target characterization~\cite{7572958}, the
EM compatibility (EMC) analysis of microwave
subsystems~\cite{ref3}, and the design and performance
evaluation of microwave sensing and radar front-ends.
For a perfectly electrically conducting (PEC) target, the
scattered field is fully determined by the induced
surface current density, which is governed by the
electric field integral equation (EFIE) obtained by
enforcing the vanishing of the tangential total electric
field on the target surface, together with the Sommerfeld
radiation condition at
infinity~\cite{harrington1996field}. Accurate and efficient numerical solution of the EFIE therefore lies at the core of CEM for microwave-frequency scattering problems.

Classical CEM solvers, including the Method of Moments
(MoM)~\cite{harrington1996field}, the Finite Element Method
(FEM)~\cite{ref2}, and the Finite-Difference Time-Domain
(FDTD) method~\cite{ref3}, have been extensively
validated across a wide range of target geometries and
frequency regimes. Among these, MoM discretizes the EFIE
with Rao--Wilton--Glisson (RWG) basis
functions~\cite{rao1982electromagnetic} and yields a
dense linear system whose impedance matrix storage scales
as $\mathcal{O}(N^{2})$ and whose direct factorization
scales as $\mathcal{O}(N^{3})$, where $N$ denotes the
number of unknowns. To mitigate this computational
burden, the Multilevel Fast Multipole Algorithm
(MLFMA)~\cite{ref5} introduces a hierarchical octree
partition that decomposes the impedance matrix into
near-field and far-field contributions, reducing the
per-iteration matrix--vector product complexity to
$\mathcal{O}(N\log N)$ and enabling the tractable
solution of electrically large microwave scattering
problems. Subsequent algorithmic developments, including
kernel compression~\cite{ref8,ref9}, fast
matrix--vector multiplication~\cite{ref10}, and
GPU-accelerated implementations~\cite{ref11,ref12}, have
further extended the scalability of MoM-based solvers to
problems involving billions of unknowns.

In recent years, deep learning has emerged as a
data-driven complement to classical CEM solvers,
offering substantial inference speedup once trained on
sufficient scattering data. Owing to their nonlinear
approximation capability, artifical neural networks can map
geometric representations of scattering targets directly
to EM responses, thereby bypassing the explicit assembly
and inversion of the impedance matrix. Representative
frameworks include supervised descent methods that infer
scattering characteristics from image-based target
descriptions~\cite{ref13}, multi-task learning
architectures that improve cross-configuration
generalization through shared latent
representations~\cite{ref14}, and models incorporating
radiation-direction maps as extended input
modalities~\cite{ref15}. At the architecture level,
fully convolutional networks have been used to predict
near-field potential distributions in volumetric EM
models~\cite{ref18}, while physics-augmented U-Net
structures impose Maxwell-consistent constraints through
auxiliary loss terms~\cite{ref19}. Despite these
advances, purely data-driven approaches tend to
underperform on geometrically complex targets and under
excitation conditions that are not well covered by the
training distribution~\cite{refre}, which limits their
applicability in physically sensitive microwave tasks
such as multi-aspect RCS computation and wideband EM
analysis.

To address the limited generalization of purely
data-driven models, physics-informed and
physics-informed CEM methods have been explored,
incorporating EM operators, Green's-function kernels,
and boundary-condition residuals into the learning
process~\cite{ref51}. For EM scattering problems,
existing physics-informed methods can be broadly grouped
into two directions: one line of work focuses on the
modeling of local interactions arising from
integral-equation discretization, while another targets
the acceleration of long-range radiation evaluation in
MLFMA-type hierarchical solvers.

At the near-field level, several studies couple neural
modules with MoM-based operators to accelerate the
solution of the integral equation while retaining
physical fidelity. Representative efforts include
iterative unrolling schemes that alternate between
current and material-parameter refinement under EFIE
constraints~\cite{ref21}, physics-regularized generative
models for inverse scattering~\cite{ref23}, and
neural-operator approximations of the integral kernel
for rapid two-dimensional field estimation~\cite{ref24}.
More recently, graph-based architectures have been
adopted to exploit the mesh topology inherent in MoM
discretizations: graph convolutional networks have been
applied to learn EFIE-structured field
interactions~\cite{Stylianopoulos2025GraphCNNsFR}, and
graph neural network modules have been embedded within
3D MoM iterative solvers to perform data-driven residual
updates on mesh-based graphs~\cite{ref28}. In a related
direction, \cite{11494458} replaces the fixed
message-passing kernels of conventional graph neural networks
with learnable univariate basis functions that embed a
Green's-function prior, yielding a learnable yet
physically grounded representation of pairwise impedance
interactions.

At the far-field level, complementary efforts have
targeted the multilevel translation operators that
dominate the cost of MLFMA. Radial basis function
networks have been employed to approximate the
translation kernel between well-separated
clusters~\cite{ref53}, and hybrid architectures
combining embedding layers with self-attention have been
proposed to jointly learn dependencies across
hierarchical octree levels~\cite{10930892}. The
physics-informed framework in~\cite{ref22} further
accelerates the translation step through a
coarse-to-fine strategy, reducing the computational cost
for electrically large targets. These results suggest
that physics-guided learning offers a viable means of
accelerating the evaluation of long-range radiation
effects that dominate far-field scattering signatures.

In this paper, we propose a U-shaped physics-informed neural 
network, termed U-PINet, for end-to-end 3D microwave
scattering modeling. The proposed framework is motivated
by repeated-query scattering analysis scenarios at
microwave frequencies, such as multi-aspect, multi-frequency,
and multi-polarization RCS evaluation, in which
conventional MLFMA solvers must be invoked under a large
number of incidence and observation configurations.
Following the near--far field decomposition inherent in
MLFMA, U-PINet integrates dedicated near-field and
far-field learning modules within a unified architecture,
and uses the EFIE boundary-condition residual as a
training objective. Once trained, the proposed network predicts
the induced surface current on a 3D PEC target in a
single forward pass, from which the scattered field and
the bistatic RCS are evaluated via the radiation integral
for arbitrary observation directions. The validation
experiments for U-PINet are conducted on canonical and
geometrically complex 3D PEC targets at microwave
frequencies in the range of $0.5$--$2$~GHz. The main contributions of this work are summarized as follows.
\begin{itemize}
    \item We propose U-PINet, a physics-informed learning framework
    for 3D microwave scattering in which the near--far decomposition
    of MLFMA, originally introduced as a numerical acceleration
    mechanism, is reused as a structural prior for neural network modeling.
    Under this prior, the induced surface current is represented as
    the sum of a short-range impedance-coupling component and a
    long-range radiation-coupling component, learned jointly within
    a unified architecture and constrained by a discretized EFIE
    residual at surface collocation points.

    \item For the near-field component, we formulate the local action
    of the MoM near-field block as a graph-supported learnable
    operator on the discretized PEC surface, where the graph connectivity is inherited from its local interaction pattern. Each pairwise
    mesh-element interaction is modeled as a function of the relative
    position, the surface-normal variation, and the neighboring
    current state, and is parameterized by learnable univariate basis
    functions. This approach yields a geometry-adaptive realization of the discretized near-field operator.

    \item For the far-field component, we design a hierarchical
    learning module for nonlocal radiation coupling, which is organized on the
    same octree partition as MLFMA. Long-range interactions are
    captured through coarsening, cross-scale information exchange,
    and coarse-to-fine recovery, in which the fixed analytic
    translation kernels of MLFMA are replaced by learnable cross-scale
    coupling, while the multi-resolution organization that makes such
    coupling tractable is retained.

    \item The resulting end-to-end differentiable surrogate is
    evaluated on seven canonical and geometrically complex 3D PEC
    targets at microwave frequencies, covering multi-frequency and
    multi-polarization current reconstruction, bistatic RCS
    prediction, and near-far ablation studies. It is demonstrated that the proposed U-PINet framework attains lower
    current-level and RCS-level errors than representative
    physics-informed baselines, with more visible gains on
    electrically larger and multi-scale geometries.
\end{itemize}

The remainder of this paper is organized as follows.
Section~\ref{section:B} presents the 3D PEC scattering formulation
and the MoM/MLFMA background. Section~\ref{sec:UPINET_framework}
introduces the functional design of U-PINet, while
Section~\ref{sec:architecture} details its network realization.
Section~\ref{section:D} describes the experimental setup, and
Section~\ref{section:E} reports the validation results, including
accuracy, ablation, solver-oriented, and runtime analyses. Finally, Section~\ref{section:conclusion} includes the concluding remarks of the paper.

\section{Problem Formulation and State of the Art}
\label{section:B} 

\subsection{Electromagnetic Scattering Modeling Formulation}
\label{subsec:formulation}

\begin{figure}[!t]
\centering
\includegraphics[width=8cm]{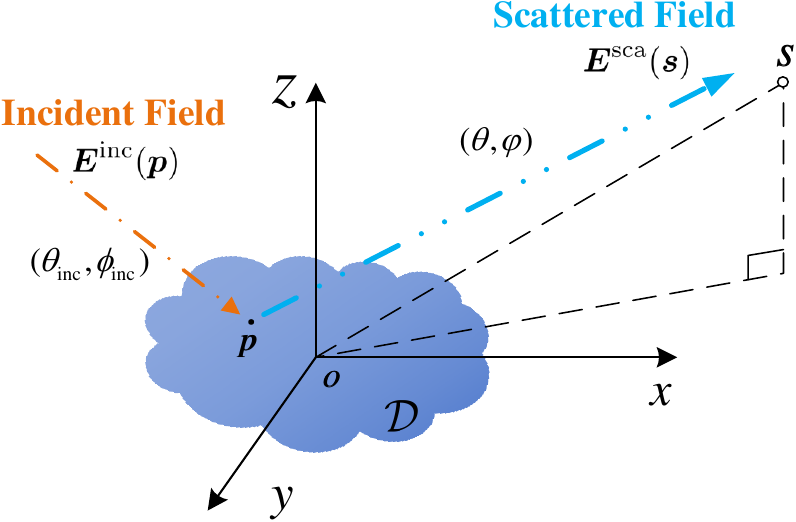}
\caption{Geometry for microwave scattering and RCS evaluation. A PEC target with surface $\mathcal{D}$ (point $\boldsymbol{p}$ belongs in this surface) is illuminated by an incident plane wave $\boldsymbol{E}^{\mathrm{inc}}$ from direction $(\theta_{\mathrm{inc}},\varphi_{\mathrm{inc}})$, and the scattered field $\boldsymbol{E}^{\mathrm{sca}}$ is observed at a far-field point $\boldsymbol{s}$ along the direction $(\theta,\varphi)$.}
\label{fig: rcs}
\end{figure}

\begin{figure*}[t]
 \centerline{\includegraphics[width=2\columnwidth,draft=false]{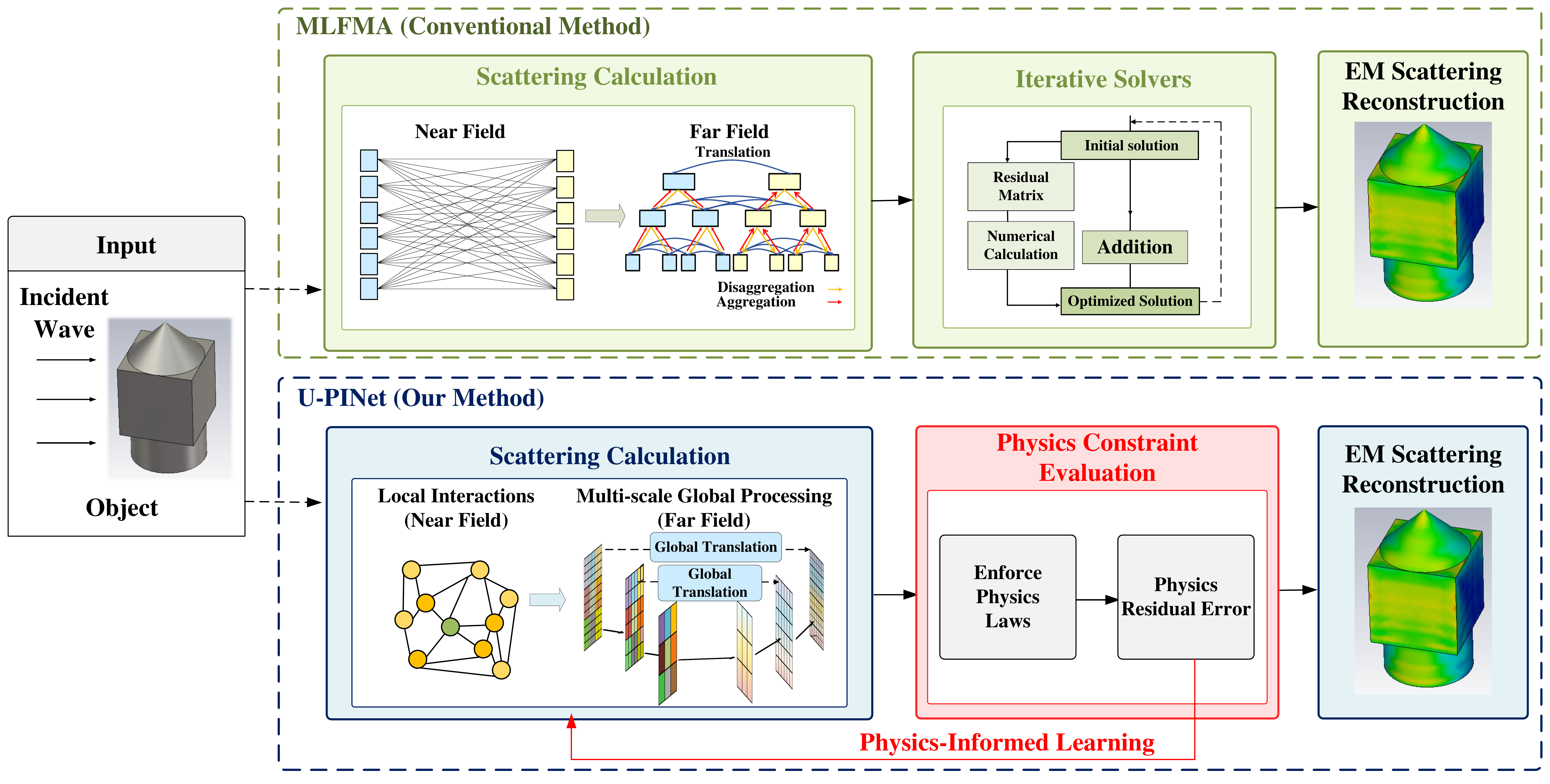}}
\caption{Microwave scattering evaluation
pipelines for the standard MLFMA algorithm~\cite{ref5}
(top) and the proposed U-PINet framework (bottom).
U-PINet retains the near--far separation of length scales
of MLFMA, but replaces the iterative linear system solver
with a single forward pass: a near-field graph branch and
a multi-scale far-field branch jointly predict the surface
current, which is regularized through an EFIE-based
physics-residual loss (see Sections~\ref{sec:UPINET_framework}
and~\ref{sec:architecture} for more details).}

 \label{fig:test1}
\end{figure*}

Consider a PEC target with closed surface $\mathcal{D}$
embedded in a homogeneous background medium, as illustrated
in Fig.~\ref{fig: rcs}. A 3D Cartesian coordinate system is
adopted with the target centered at the origin
$\boldsymbol{o}$, and a time-harmonic
$e^{j\omega t}$ convention is assumed throughout this paper.
The target is illuminated by an incident plane wave
$\boldsymbol{E}^{\mathrm{inc}}(\boldsymbol{p})$ propagating
along the direction
$(\theta_{\mathrm{inc}}, \varphi_{\mathrm{inc}})$, which
induces a tangential surface current density
$\boldsymbol{J}(\boldsymbol{p})$ on $\mathcal{D}$. The
induced current acts as the secondary radiation source for
the scattered field
$\boldsymbol{E}^{\mathrm{sca}}$ in the exterior region, so
the central physical task in EM scattering analysis is the
reconstruction of $\boldsymbol{J}$.

Expressing the scattered field in terms of $\boldsymbol{J}$ through 
the dyadic Green's function of the background medium and imposing 
the Sommerfeld radiation condition at infinity leads to the EFIE:
\begin{equation}
    \hat{\boldsymbol{n}}(\boldsymbol{p}) \times
    \left(
        \boldsymbol{E}^{\mathrm{inc}}(\boldsymbol{p})
        +
        \int_{\mathcal{D}}
            \mathbf{G}(\boldsymbol{p}, \boldsymbol{p}')
            \cdot
            \boldsymbol{J}(\boldsymbol{p}')\,
            \mathrm{d}S'
    \right)
    = \boldsymbol{0},
    \quad \boldsymbol{p} \in \mathcal{D},
    \label{eq:efie_operator}
\end{equation}
where ``$\times$'' denotes the vector cross product, used here 
to extract the tangential component of the bracketed field on 
$\mathcal{D}$; ``$\cdot$'' denotes the contraction of the dyadic 
kernel $\mathbf{G}$ with the vector current $\boldsymbol{J}$, 
yielding the vector electric field radiated by $\boldsymbol{J}$. 
The scattered-field integral is understood as the limit taken 
from the exterior side of $\mathcal{D}$, and 
$\mathbf{G}(\boldsymbol{p}, \boldsymbol{p}')$ denotes the dyadic 
kernel relating a surface current at $\boldsymbol{p}'$ to the 
resulting electric field at $\boldsymbol{p}$...

Once $\boldsymbol{J}$ is reconstructed, the scattered
field at an exterior observation point $\boldsymbol{s}$
parameterized by the spherical angles $(\theta,\varphi)$ is
obtained from the radiation integral
\begin{equation}
    \boldsymbol{E}^{\mathrm{sca}}(\boldsymbol{s})
    =
    \int_{\mathcal{D}}
        \mathbf{G}(\boldsymbol{s},\boldsymbol{p})
        \cdot
        \boldsymbol{J}(\boldsymbol{p})\,
        \mathrm{d}S(\boldsymbol{p}),
    \label{eq:Esca_operator}
\end{equation}
in which $\mathbf{G}(\boldsymbol{s},\boldsymbol{p})$ denotes
the same dyadic Green's function as in
\eqref{eq:efie_operator}, now evaluated between a surface
source point $\boldsymbol{p}\in\mathcal{D}$ and an exterior
observation point $\boldsymbol{s}$ lying outside
$\mathcal{D}$.In the asymptotic limit $\|\boldsymbol{s}\|\to\infty$ of large
observation distance, $\mathbf{G}(\boldsymbol{s},\boldsymbol{p})$
admits the standard radiation-zone form that retains the
outgoing spherical-wave factor and the directional pattern along
$(\theta,\varphi)$, so that~\eqref{eq:Esca_operator} can be
evaluated for arbitrary observation directions without having to
re-solve the EFIE in~\eqref{eq:efie_operator}.

The corresponding bistatic RCS along the
direction~$(\theta,\varphi)$ is defined as follows~\cite{balanis2012advanced}:
\begin{equation}
    \sigma(\theta,\varphi) = \lim_{r\to\infty} 4\pi r^2 \,
\frac{\|\boldsymbol{E}^{\mathrm{sca}}(r,\theta,\varphi)\|^2}
{\|\boldsymbol{E}^{\mathrm{inc}}\|^2},
    \label{eq:rcs_def}
\end{equation}
where $r = \|\boldsymbol{s}\|$ is the distance from the target to the
observation point, $\|\cdot\|$ denotes the Euclidean norm of a
complex-valued electric-field vector, defined through 
$\|\boldsymbol{E}\|^2 = \boldsymbol{E} \cdot \boldsymbol{E}^*$, 
in which ``$\cdot$'' denotes the standard vector inner product 
between two complex $3$-vectors (with no Hermitian conjugation 
absorbed into the operator itself), and $(\cdot)^*$ denotes 
elementwise complex conjugation, so that 
$\boldsymbol{E} \cdot \boldsymbol{E}^* 
= \sum_{i} E_i E_i^* = \sum_i |E_i|^2$ is real and non-negative.

Although $\sigma(\theta,\varphi)$ provides a compact
scalar descriptor at a specific observation direction,
the reconstructed current~$\boldsymbol{J}$ carries the
full physical information of the scattering response and
fully determines the scattered field via
\eqref{eq:Esca_operator}. The proposed framework
therefore targets the high-fidelity reconstruction
of~$\boldsymbol{J}$, from which both monostatic and
bistatic RCS can be evaluated as downstream validation
metrics through the previous expressions
\eqref{eq:Esca_operator} and \eqref{eq:rcs_def}.

\subsection{The MLFMA Framework}
\label{subsec:mlfma}

As established in Section~\ref{subsec:formulation}, the
central task in EM scattering analysis is to determine
the induced surface current density $\boldsymbol{J}$ from
the EFIE in~\eqref{eq:efie_operator}. The standard
numerical approach discretizes~\eqref{eq:efie_operator}
through the Method of Moments with RWG basis functions $\{\boldsymbol{f}_n\}_{n=1}^{N}$
defined on the edges of a triangular surface
mesh~\cite{rao1982electromagnetic}. Under the Galerkin scheme, the unknown
surface current is expanded as
$\boldsymbol{J}(\boldsymbol{p}) = \sum_{n=1}^{N} I_n\,
\boldsymbol{f}_n(\boldsymbol{p})$, and the integral
equation is tested with the same set of functions,
yielding the dense linear system:
\begin{equation}
    \mathbf{Z}\,\boldsymbol{I} = \boldsymbol{V},
    \label{eq:mom_system}
\end{equation}
where $\mathbf{Z}\in\mathbb{C}^{N\times N}$ is the
impedance matrix, $\boldsymbol{I}\in\mathbb{C}^{N}$
collects the expansion coefficients, and
$\boldsymbol{V}\in\mathbb{C}^{N}$ contains the tested
incident field. Each entry of $\mathbf{Z}$ is given by
the Galerkin inner product, as follows:
\begin{equation}
    Z_{mn}
    = \int_{\mathcal{D}} \boldsymbol{f}_m(\boldsymbol{p}) \cdot
      \int_{\mathcal{D}} \mathbf{G}(\boldsymbol{p},\boldsymbol{p}')
      \cdot \boldsymbol{f}_n(\boldsymbol{p}')\,
      \mathrm{d}S'\, \mathrm{d}S,
    \label{eq:zmn_entry}
\end{equation}
where the inner integral evaluates the field radiated by
$\boldsymbol{f}_n$ through $\mathbf{G}$, and the outer
integral tests this field against $\boldsymbol{f}_m$. The
kernel $\mathbf{G}(\boldsymbol{p},\boldsymbol{p}')$ is
taken in its mixed-potential form, so that
\eqref{eq:zmn_entry} aggregates both the vector- and
scalar-potential contributions of the EFIE operator,
with the surface divergence transferred onto the RWG
basis and testing functions and near-singular and
self-term integrals handled by standard
singularity-extraction techniques. Direct evaluation of
the matrix--vector product $\mathbf{Z}\boldsymbol{I}$
requires $\mathcal{O}(N^{2})$ operations, which becomes
prohibitive for electrically large targets.

To mitigate this cost, the MLFMA~\cite{ref5} decomposes
the impedance matrix into near-field and far-field
contributions,
\begin{equation}
    \mathbf{Z}
    = \mathbf{Z}_{\mathrm{near}}
    + \mathbf{Z}_{\mathrm{far}},
    \label{eq:z_decomp}
\end{equation}
and treats the two parts with distinct computational
strategies, as illustrated in Fig.~\ref{fig:test1}. The
overall algorithm proceeds in three stages: exact
evaluation of the near-field block, hierarchical
evaluation of the far-field block, and residual-based
iterative solution of the resulting linear system.

For each pair of basis functions $(m,n)$ whose supports
are geometrically close, the impedance entry $Z_{mn}$
in~\eqref{eq:zmn_entry} is evaluated exactly and stored
in the sparse block $\mathbf{Z}_{\mathrm{near}}$, so that
the product $\mathbf{Z}_{\mathrm{near}}\boldsymbol{I}$
accounts for all local self- and mutual-coupling
contributions. The accuracy of this block is governed by
the basis functions $\{\boldsymbol{f}_n\}$, which jointly
determine the spatial resolution of the current
representation and the fidelity of the local impedance
interactions.

The remaining long-range interactions are organized on an
$L$-level octree~\cite{octree}, where mesh elements are
recursively grouped into well-separated source and
observation clusters. At each level $l$, the radiation pattern
$\boldsymbol{\beta}^{(l)}$ is sampled on a set of
directions $\{\hat{\boldsymbol{k}}_q^{(l)}\}$ on the unit
sphere, and the translation operator
$\mathcal{T}_l(\hat{\boldsymbol{k}})$ acts diagonally in
$\hat{\boldsymbol{k}}$, so that the multipole translation
step amounts to a pointwise multiplication followed by a
summation over the interaction list.
 Well-separated interactions are then handled by a
level-dependent translation operator $\mathcal{T}_l$,
which maps the aggregated patterns
$\boldsymbol{\beta}^{(l)}$ of groups in the interaction
list $\mathcal{I}$ to incoming fields
$\boldsymbol{\alpha}_{\mathrm{trans}}^{(l)}$. The
translated incoming fields are subsequently disaggregated
from coarse to fine levels. At the finest level, the
accumulated field $\boldsymbol{\alpha}^{(L)}$ yields the
far-field contribution
$\mathbf{Z}_{\mathrm{far}}\boldsymbol{I}$, recovering the
full matrix--vector product as
$\mathbf{Z}\boldsymbol{I}
= \mathbf{Z}_{\mathrm{near}}\boldsymbol{I}
+ \mathbf{Z}_{\mathrm{far}}\boldsymbol{I}$ with an overall
per-iteration cost of $\mathcal{O}(N\log N)$.

The product $\mathbf{Z}\boldsymbol{I}$ is then embedded
in a Krylov subspace solver such as GMRES~\cite{gmres} or
BiCGStab~\cite{bicg}, which iteratively updates
$\boldsymbol{I}$ by reducing the residual
$\boldsymbol{r} = \boldsymbol{V} -
\mathbf{Z}\boldsymbol{I}$ until a prescribed convergence
tolerance is reached. Upon convergence, the coefficient
vector $\boldsymbol{I}$ is substituted back into the
basis-function expansion to reconstruct
$\boldsymbol{J}(\boldsymbol{p})
= \sum_n I_n\,\boldsymbol{f}_n(\boldsymbol{p})$, from
which the scattered field and the bistatic RCS are
obtained via~\eqref{eq:Esca_operator}
and~\eqref{eq:rcs_def}, respectively.
Algorithm~\ref{alg:mlfma_scheme} summarizes the
resulting near--far computational pattern, which serves
as the reference structure for the physics-informed
learning architecture developed in
Section~\ref{sec:UPINET_framework}.

\begin{algorithm}[t!]
\caption{The State-of-the-Art MLFMA Procedure}
\label{alg:mlfma_scheme}
\begin{algorithmic}[1]
\Require
    RWG basis functions $\{\boldsymbol{f}_n\}_{n=1}^{N}$;
    number of octree levels~$L$;
    translation operators~$\{\mathcal{T}_l\}$;
    tested incident field~$\boldsymbol{V}$;
    observation angles~$(\theta,\varphi)$.
\Ensure
    Induced surface current~$\boldsymbol{J}$ and bistatic RCS~$\sigma(\theta,\varphi)$.

\Statex \textbf{Step 1: Octree Partition and Neighbor Lists}
\State Partition the mesh hierarchically into $L$ octree levels, from level~$1$ (root) to level~$L$ (leaves).
\State For each octree group, build the near-neighbor list~$\mathcal{N}$ and the interaction list~$\mathcal{I}$.

\vspace{0.3em}
\Statex \textbf{Step 2: Near-Field Block Assembly}
\For{each near-field pair $(m,n) \in \mathcal{N}$}
    \State Evaluate $Z_{mn}$ exactly via~\eqref{eq:zmn_entry} using $\boldsymbol{f}_m$ and~$\boldsymbol{f}_n$.
\EndFor
\State Assemble the sparse near-field block~$\mathbf{Z}_{\mathrm{near}}$ and compute~$\mathbf{Z}_{\mathrm{near}}\boldsymbol{I}$.

\vspace{0.3em}
\Statex \textbf{Step 3: Far-Field Aggregation}
\State Initialize the leaf-level radiation patterns:
       $\boldsymbol{\beta}^{(L)} \leftarrow \mathcal{F}(\boldsymbol{I})$.
\For{$l = L$ down to $2$}
    \State $\boldsymbol\beta^{(l-1)} \leftarrow \sum_{\mathrm{child}} \mathrm{Interp}\!\left(\boldsymbol\beta^{(l)} \cdot e^{j\bm k \cdot \bm d}\right)$
\EndFor

\vspace{0.3em}
\Statex \textbf{Step 4: Multipole Translation}
\For{$l = 2$ to $L$}
    \State $\boldsymbol{\alpha}^{(l)}_{\mathrm{trans},\,i} \leftarrow
           \sum_{m \in \mathcal{I}(i)} \mathcal{T}_{l}\,
           \boldsymbol{\beta}^{(l)}_m$ for every observation group~$i$ at level~$l$.
\EndFor

\vspace{0.3em}
\Statex \textbf{Step 5: Far-Field Disaggregation}
\State Initialize $\boldsymbol{\alpha}^{(1)} \leftarrow \boldsymbol{0}$.
\For{$l = 2$ to $L$}
    \State $\boldsymbol\alpha^{(l)} \leftarrow \mathrm{Anterp}\!\left(\boldsymbol\alpha^{(l-1)} \cdot e^{-j\bm k \cdot \bm d}\right) + \boldsymbol\alpha^{(l)}_{\mathrm{trans}}$
\EndFor
\State Project onto testing functions:
       $\mathbf{Z}_{\mathrm{far}}\boldsymbol{I} \leftarrow
       \int \boldsymbol{\alpha}^{(L)}(\hat{\boldsymbol{k}}) \cdot
       \boldsymbol{f}(\hat{\boldsymbol{k}})\, \mathrm{d}\hat{\boldsymbol{k}}$.

\vspace{0.3em}
\Statex \textbf{Step 6: Iterative Linear System Solution}
\State Combine the two contributions:
       $\mathbf{Z}\boldsymbol{I} \leftarrow
       \mathbf{Z}_{\mathrm{near}}\boldsymbol{I}
       + \mathbf{Z}_{\mathrm{far}}\boldsymbol{I}$.
\State Compute the residual
       $\boldsymbol{r} \leftarrow \boldsymbol{V} - \mathbf{Z}\boldsymbol{I}$.
\State Update~$\boldsymbol{I}$ via GMRES or BiCGStab to reduce~$\|\boldsymbol{r}\|$ to a prescribed tolerance.

\vspace{0.3em}
\Statex \textbf{Step 7: Current Recovery and RCS Evaluation}
\State Reconstruct $\boldsymbol{J}(\boldsymbol{p})
       = \sum_{n=1}^{N} I_n\,\boldsymbol{f}_n(\boldsymbol{p})$ from the converged~$\boldsymbol{I}$.
\State Evaluate $\boldsymbol{E}^{\mathrm{sca}}$ via~\eqref{eq:Esca_operator}
       and obtain $\sigma(\theta,\varphi)$ from~\eqref{eq:rcs_def}.
\end{algorithmic}
\end{algorithm}

\begin{algorithm}[t!]
\caption{The Proposed U-PINet Procedure}
\label{alg:upinet_mlfma_style}
\begin{algorithmic}[1]
\Require
    Surface points $\mathcal{P} = \{\boldsymbol{p}_i\}$ with normals~$\{\hat{\boldsymbol{n}}_i\}$;
    near-field radius~$r_{\mathrm{nf}}$;
    number of hierarchical levels~$L$;
    number of message-passing iterations~$T_{\mathrm{iter}}$;
    batch of training incident
    fields~$\{\boldsymbol{E}^{\mathrm{inc},(m)}\}_{m=1}^{N_s}$;
    query incident field~$\boldsymbol{E}^{\mathrm{inc}}$;
    observation angles~$(\theta,\varphi)$.
\Ensure
    Trained parameters~$\Theta^{\ast}$, predicted surface
    current~$\boldsymbol{J}_{\mathrm{pred}}$, and bistatic
    RCS~$\sigma(\theta,\varphi)$.

\Statex \textbf{Step 1: Near-Field Graph Construction}
\For{each $\boldsymbol{p}_i \in \mathcal{P}$}
    \State $\mathcal{N}(i) \leftarrow
           \{\,j : \|\boldsymbol{p}_i - \boldsymbol{p}_j\| \le r_{\mathrm{nf}}\,\}$
           \hfill via~\eqref{eq:nf_nbr}
    \State $\boldsymbol{w}_{ij} \leftarrow
           \mathcal{K}\!\big(\|\boldsymbol{p}_i - \boldsymbol{p}_j\|,\;
           \hat{\boldsymbol{n}}_i - \hat{\boldsymbol{n}}_j\big)$
           \hfill via~\eqref{eq:edge_feat}
\EndFor

\vspace{0.3em}
\Statex \textbf{Step 2: Iterative Near-Field Message Passing}
\State Initialize $\boldsymbol{J}^{(0)}(\boldsymbol{p}_i) \leftarrow \boldsymbol{0}$
       for all~$\boldsymbol{p}_i \in \mathcal{P}$.
\For{$t = 0$ to $T_{\mathrm{iter}} - 1$}
    \For{each $\boldsymbol{p}_i \in \mathcal{P}$}
        \State Form edge descriptors
               $\{\boldsymbol{\xi}_{ij}^{(t)}\}_{j\in\mathcal{N}(i)}$
               \hfill via~\eqref{eq:edge_descriptor}
        \State Compute edge messages
               $\boldsymbol{m}_{ij}^{(t)} \leftarrow
               \Phi^{(t)}\!\big(\boldsymbol{\xi}_{ij}^{(t)}\big)$
               \hfill via~\eqref{eq:edge_message}
        \State Aggregate and update with edge-attention weights:
       $\boldsymbol{J}^{(t+1)}(\boldsymbol{p}_i) \leftarrow
       \boldsymbol{J}^{(t)}(\boldsymbol{p}_i) +
       \sum_{j\in\mathcal{N}(i)} \rho_{ij}\,\boldsymbol{m}_{ij}^{(t)}$,
       where $\rho_{ij}$ is the softmax edge weight
       in~\eqref{eq:gkan_aggregation}.
       \hfill via~\eqref{eq:gnn_update}
    \EndFor
\EndFor
\State $\boldsymbol{J}_{\mathrm{near}} \leftarrow
       \boldsymbol{J}^{(T_{\mathrm{iter}})}$

\vspace{0.3em}
\Statex \textbf{Step 3: Hierarchical Feature Pooling}
\State $\boldsymbol{S}^{(L)} \leftarrow \boldsymbol{J}_{\mathrm{near}}$
\For{$k = L$ down to $2$}
    \State $\boldsymbol{S}^{(k-1)} \leftarrow
           \boldsymbol{\Gamma}_k\!\big(\boldsymbol{S}^{(k)}\big)$
           \hfill via~\eqref{eq:upinet_P}
\EndFor

\vspace{0.3em}
\Statex \textbf{Step 4: Cross-Scale Feature Exchange}
\State $\tilde{\boldsymbol{S}}^{(1)} \leftarrow \boldsymbol{S}^{(1)}$
\For{$k = 2$ to $L$}
    \State $\tilde{\boldsymbol{S}}^{(k)} \leftarrow
           \boldsymbol{\mathcal{A}}_k\!\big(
           \boldsymbol{S}^{(k)},\;
           \mathcal{U}_k(\tilde{\boldsymbol{S}}^{(k-1)})
           \big)$
           \hfill via~\eqref{eq:upinet_Phi_def}
\EndFor

\vspace{0.3em}
\Statex \textbf{Step 5: Coarse-to-Fine Feature Fusion}
\State $\boldsymbol{J}_{\mathrm{far}}^{(1)} \leftarrow \tilde{\boldsymbol{S}}^{(1)}$
\For{$k = 2$ to $L$}
    \State $\boldsymbol{J}_{\mathrm{far}}^{(k)} \leftarrow
           \boldsymbol{\Psi}_k\!\big(
           \boldsymbol{J}_{\mathrm{far}}^{(k-1)},\;
           \tilde{\boldsymbol{S}}^{(k)}\big)$
           \hfill via~\eqref{eq:upinet_disagg}
\EndFor
\State $\boldsymbol{J}_{\mathrm{far}} \leftarrow
       \boldsymbol{J}_{\mathrm{far}}^{(L)}$

\vspace{0.3em}
\Statex \textbf{Step 6: Physics-Informed Training Update}
\State Assemble
       $\boldsymbol{J}_{\mathrm{pred}} \leftarrow
       \boldsymbol{J}_{\mathrm{near}} + \boldsymbol{J}_{\mathrm{far}}$
       \hfill via~\eqref{eq:Jpred_assemble}
\State Compute $\mathcal{L}_{\mathrm{phys}}$ over the training
       batch and update $\Theta$ by gradient-based optimization
       \hfill via~\eqref{eq:physics_loss}

\vspace{0.3em}
\Statex \textbf{Step 7: Inference and RCS Evaluation}
\State With $\Theta^{\ast}$, predict
       $\boldsymbol{J}_{\mathrm{pred}}$ for a query incident field
       by one forward pass through Steps~1--5.
\State Evaluate $\boldsymbol{E}^{\mathrm{sca}}(\boldsymbol{s})$
       and compute $\sigma(\theta,\varphi)$
       \hfill via~\eqref{eq:Esca_operator} and~\eqref{eq:rcs_def}
\end{algorithmic}
\end{algorithm}

\section{The U-PINet Neural Network Framework}
\label{sec:UPINET_framework}

The design of the proposed U-PINet is based on a simple physical principle: EM scattering of microwave-frequency plane waves from a $3$D PEC target, represented by a discretized surface mesh, contains interaction patterns at different spatial scales, and
these patterns should not be represented by a single uniform
neural mapping. Local impedance coupling is mainly determined by
nearby mesh elements and their geometric relationship, whereas
long-range radiation coupling is mediated by distant surface
regions and has an inherent multi-scale structure. U-PINet capitalizes upon this scale-separated interaction structure as the main organizing
principle for its architecture. From this perspective, the role of MLFMA in this work is not to provide an operator that is directly reproduced by the network, instead, this algorithm is deployed to provide a physically meaningful computational
organization: local interactions are treated in a neighborhood
form, while nonlocal interactions are represented through a
hierarchical grouping of the surface unknowns. We reinterpret this
organization as a neural network modeling principle for surface-current
prediction. The resulting architecture is, therefore, designed to
match the physical structure of the underlying scattering process, instead
of using a generic neural network to learn all interactions in the same
latent space.

As shown in Fig.~\ref{fig:test1}, the aforedescribed U-PINet principle leads to a
two-branch architecture. A near-field graph branch operates on a
local neighborhood induced by the surface mesh, and produces a
short-range estimate $\boldsymbol{J}_{\mathrm{near}}$ of the
induced current (Section~\ref{subsec:nearfield_func}).In addition, a
hierarchical far-field branch operates on a multi-resolution
representation of the surface, and produces a complementary
long-range contribution $\boldsymbol{J}_{\mathrm{far}}$
(Section~\ref{subsec:farfield_func}). The two contributions are then
summed to form the predicted surface current, as follows:
\begin{equation}
    \boldsymbol{J}_{\mathrm{pred}}
    =
    \boldsymbol{J}_{\mathrm{near}}
    +
    \boldsymbol{J}_{\mathrm{far}}.
    \label{eq:Jpred_assemble}
\end{equation}
This current is regularized during training through an EFIE-based
residual loss that promotes consistency with the PEC boundary
condition in~\eqref{eq:efie_operator}
(Section~\ref{subsec:physics_loss}). After training,
$\boldsymbol{J}_{\mathrm{pred}}$ is substituted into the
radiation integral in~\eqref{eq:Esca_operator} to evaluate the
scattered field, and the bistatic
RCS~$\sigma(\theta,\varphi)$ is obtained
from expression ~\eqref{eq:rcs_def} for arbitrary observation directions.

The correspondence between the classical MLFMA pipeline in
Algorithm~\ref{alg:mlfma_scheme} and the proposed learning
procedure in Algorithm~\ref{alg:upinet_mlfma_style} is therefore
conceptual rather than operator-equivalent. U-PINet keeps the
near--far separation of interaction scales and the hierarchical
organization of the surface unknowns, while the actual
computations are realized by trainable graph and attention-based
modules. The network-level realization of these modules is
presented in Section~\ref{sec:architecture}.

\subsection{Near-Field Interaction Learning}
\label{subsec:nearfield_func}

The near-field branch implements the local part of the
operator-decomposition principle described above. In the discretized
EFIE, the near-field block represents impedance interactions between
geometrically close basis or testing functions. These interactions are
local, geometry-sensitive, and naturally supported on the surface mesh.
We therefore model this component as a learnable graph operator
defined on the discretized surface point set
$\mathcal{P}=\{\boldsymbol{p}_i\}$, with
$\mathcal{G}=(\mathcal{P},\mathcal{E})$ denoting the resulting graph.

Following the
observation that short-range impedance coupling
in~\eqref{eq:zmn_entry} decays rapidly with distance, the
neighborhood of node~$\boldsymbol{p}_i$ is restricted to a
local radius:
\begin{equation}
    \mathcal{N}(i)
    =
    \left\{ j :
    \big\|\boldsymbol{p}_i - \boldsymbol{p}_j\big\|
    \le r_{\mathrm{nf}} \right\},
    \label{eq:nf_nbr}
\end{equation}
where $r_{\mathrm{nf}}$ is chosen proportional to the
local electrical size, which provides a physically
interpretable criterion for selecting near neighbors. On
each edge~$(i,j)\in\mathcal{E}$, an edge feature is
computed as follows:
\begin{equation}
    \boldsymbol{w}_{ij}
    =
    \mathcal{K}\!\Big(
        \big\|\boldsymbol{p}_i - \boldsymbol{p}_j\big\|,\;
        \hat{\boldsymbol{n}}_i - \hat{\boldsymbol{n}}_j
    \Big),
    \label{eq:edge_feat}
\end{equation}
where $\mathcal{K}(\cdot)$ is a learnable function whose
two arguments, inter-point distance and surface-normal
mismatch, are the same geometric quantities that govern
the magnitude of the impedance entries
in~\eqref{eq:zmn_entry}.

The induced current at each node is then refined through
iterative message passing on $\mathcal{K}$. At
iteration~$t$, a composite edge descriptor:
\begin{equation}
    \boldsymbol{\xi}_{ij}^{(t)}
    =
    \Big[
        \Delta\boldsymbol{p}_{ij},\;
        \Delta\hat{\boldsymbol{n}}_{ij},\;
        w_{ij},\;
        \boldsymbol{J}^{(t)}(\boldsymbol{p}_j)
    \Big],
    \label{eq:edge_descriptor}
\end{equation}
with
$\Delta\boldsymbol{p}_{ij}=\boldsymbol{p}_i-\boldsymbol{p}_j$
and
$\Delta\hat{\boldsymbol{n}}_{ij}=\hat{\boldsymbol{n}}_i-\hat{\boldsymbol{n}}_j$,
encodes the relative geometry between
neighbors~$\boldsymbol{p}_i,\boldsymbol{p}_j$ together
with the current state at~$\boldsymbol{p}_j$. Then, a learnable
message kernel~$\Phi^{(t)}$ maps each edge descriptor to
an edge message, as follows:
\begin{equation}
    \boldsymbol{m}_{ij}^{(t)}
    = \Phi^{(t)}\!\big(\boldsymbol{\xi}_{ij}^{(t)}\big),
    \label{eq:edge_message}
\end{equation}
and the node update takes the additive message-passing form:
\begin{equation}
    \boldsymbol{J}^{(t+1)}(\boldsymbol{p}_i)
    =
    \boldsymbol{J}^{(t)}(\boldsymbol{p}_i)
    +
    \sum_{j\in\mathcal{N}(i)}
        \boldsymbol{m}_{ij}^{(t)} .
    \label{eq:gnn_update}
\end{equation}
The summation in the latter recursion parallels the algebraic
structure of the MoM near-field action
$[\mathbf{Z}_{\mathrm{near}}\mathbf{I}]_i =
\sum_{j\in\mathcal{N}(i)} Z_{ij} I_j$,
with $\Phi^{(t)}$ playing the role of a learnable local
impedance operator on the surface graph. In the network
realization, the messages $\{\boldsymbol{m}_{ij}^{(t)}\}$
are further combined through edge-dependent attention
weights derived from the physics-guided edge kernel
$\mathcal{K}(\cdot)$ in~\eqref{eq:edge_feat},
yielding the softmax-weighted update specified
in~\eqref{eq:gkan_aggregation}. The specific network realization of $\mathcal{K}(\cdot)$ and
$\Phi^{(t)}$, including the basis-function parameterization
motivated by the Kolmogorov--Arnold representation
theorem~\cite{kan_theorem,ref41},
is presented in Section~\ref{subsec:nf_encoder}.

After $T_{\mathrm{iter}}$ message-passing iterations, the
near-field branch produces the short-range component
$\boldsymbol{J}_{\mathrm{near}}
=\boldsymbol{J}^{(T_{\mathrm{iter}})}$, which is forwarded
to the far-field branch described next.

\begin{figure*}[t]
 \centerline{\includegraphics[width=2\columnwidth,draft=false]{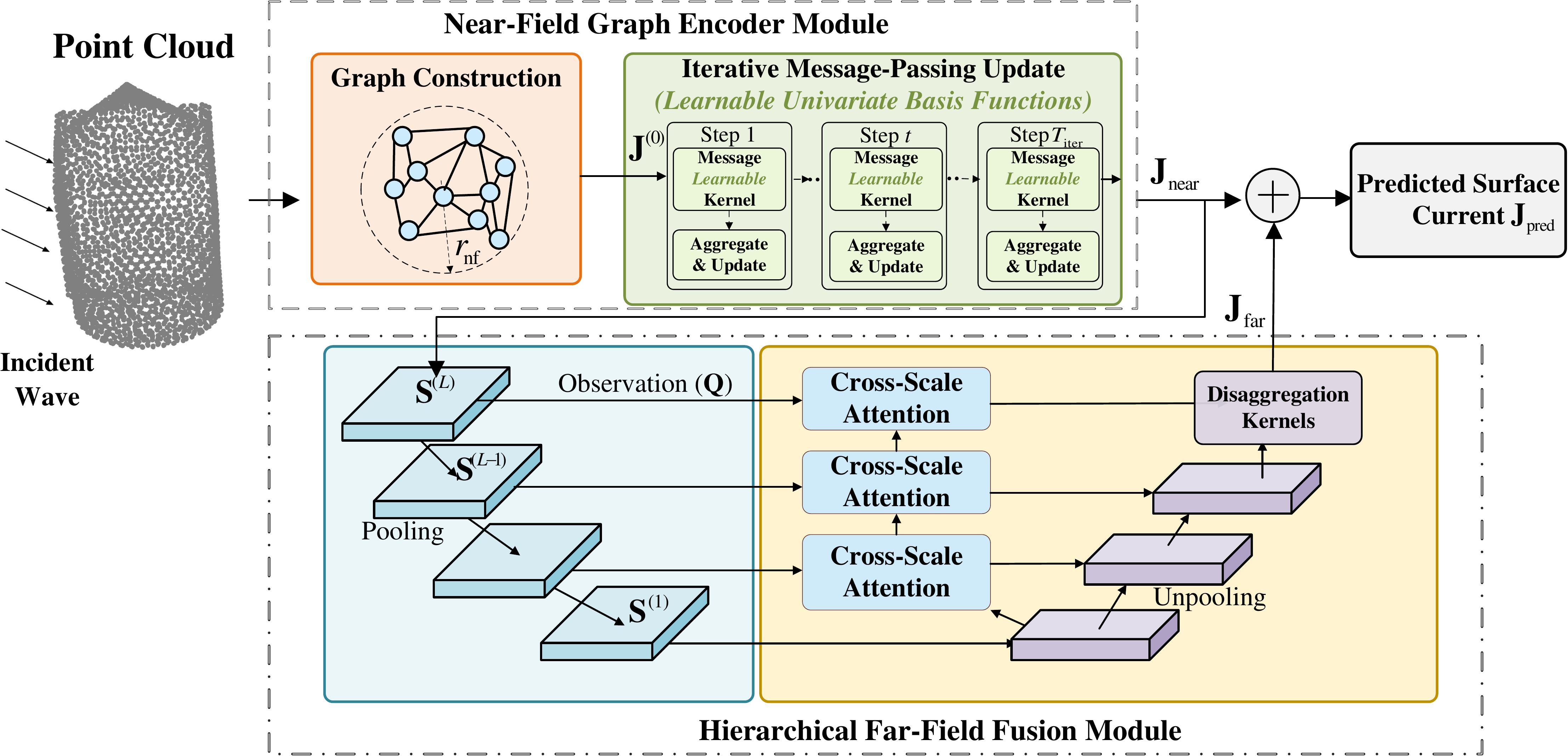}}
\caption{Architecture of the proposed U-PINet. The
near-field graph encoder  builds a
local graph on the surface mesh and applies iterative
message passing with learnable univariate basis functions
to produce~$\boldsymbol{J}_{\mathrm{near}}$. The
hierarchical far-field fusion module operates on an octree partition with pooling,
cross-scale attention, and unpooling/disaggregation to
produce~$\boldsymbol{J}_{\mathrm{far}}$. The two
contributions are summed into the predicted surface
current~$\boldsymbol{J}_{\mathrm{pred}}$ used in the
physics-residual loss.}

 \label{fig:test2}
\end{figure*}

\subsection{Far-Field Interaction Learning}
\label{subsec:farfield_func}

The far-field branch implements the nonlocal part of the same operator-decomposition principle. Unlike the near-field component, long-range radiation coupling is not well represented by purely local message passing, since distant surface regions can contribute coherently to the induced current and the scattered field. Following the multilevel organization of MLFMA, we model this component through
a hierarchical operator defined on an octree partition of the surface
points. Let
$\boldsymbol{S}^{(k)}$ denote the feature representation at
octree level~$k$, with $k=L$ corresponding to the finest
level and $k=1$ to the coarsest. The branch consists of
three functional operators that act on this multi-level
representation: \textit{i}) a pooling
operator~$\boldsymbol{\Gamma}_k$; \textit{ii}) a cross-scale coupling
operator~$\boldsymbol{\mathcal{A}}_k$, and \textit{iii}) a coarse-to-fine fusion
operator~$\boldsymbol{\Psi}_k$.

\subsubsection{Hierarchical Pooling} Starting from
$\boldsymbol{S}^{(L)}=\boldsymbol{J}_{\mathrm{near}}$,
coarser-level features are obtained by a learnable
pooling operator
\begin{equation}
    \boldsymbol{S}^{(k-1)}
    = \boldsymbol{\Gamma}_k\!\big(\boldsymbol{S}^{(k)}\big),
    \quad k = L,\dots,2,
    \label{eq:upinet_P}
\end{equation}
which produces a compact summary of the source features
within each parent cell.

\subsubsection{Cross-scale feature exchange} Long-range
interactions are modeled by a cross-scale coupling
operator that lets the fine-scale features at level~$k$
attend to a coarser context lifted from the previous
level:
\begin{equation}
    \tilde{\boldsymbol{S}}^{(k)}
    = \boldsymbol{\mathcal{A}}_k\!\big(
        \boldsymbol{S}^{(k)},\;
        \mathcal{U}_k(\tilde{\boldsymbol{S}}^{(k-1)})
      \big),
    \quad k = 2,\dots,L,
    \label{eq:upinet_Phi_def}
\end{equation}
where $\mathcal{U}_k(\cdot)$ is an unpooling operator that
lifts the coarse-level representation to level~$k$. The
operator~$\boldsymbol{\mathcal{A}}_k$ is data-dependent: it
selectively weights the coarser-scale context for each
fine-scale element, which is the mechanism through which
the branch captures long-range coupling between distant
parts of the surface.

\subsubsection{Coarse-to-Fine Fusion} The far-field current at
level~$k$ is obtained by fusing the cross-scale
output~$\tilde{\boldsymbol{S}}^{(k)}$ with the upsampled
far-field current from the coarser level:
\begin{equation}
    \boldsymbol{J}_{\mathrm{far}}^{(k)}
    =
    \boldsymbol{\Psi}_k\!\big(
        \boldsymbol{J}_{\mathrm{far}}^{(k-1)},\;
        \tilde{\boldsymbol{S}}^{(k)}
    \big),
    \quad k = 2,\dots,L.
    \label{eq:upinet_disagg}
\end{equation}
At the finest level, the far-field branch outputs
$\boldsymbol{J}_{\mathrm{far}}
= \boldsymbol{J}_{\mathrm{far}}^{(L)}$, which is combined
with~$\boldsymbol{J}_{\mathrm{near}}$ into the predicted
current $\boldsymbol{J}_{\mathrm{pred}}$
in expression~\eqref{eq:Jpred_assemble}.

The three operators
in~\eqref{eq:upinet_P}--\eqref{eq:upinet_disagg} share the
same coarse-to-fine, multi-resolution organization as
MLFMA, in the sense that distant interactions are
mediated through coarser-scale representations rather than
evaluated pairwise. The operators themselves, however, are
standard neural network components and are not intended as
numerical approximations of the multipole aggregation,
translation, or disaggregation operators of MLFMA. The
network-level realization
of~$\boldsymbol{\Gamma}_k$, $\boldsymbol{\mathcal{A}}_k$,
and~$\boldsymbol{\Psi}_k$ is presented in
Section~\ref{sec:far_field_module}.

\subsection{Physics-Informed Residual Enforcement}
\label{subsec:physics_loss}

U-PINet is trained directly against the discretized EFIE
residual, without using reference current labels generated
by a numerical solver. For each incident
field~$\boldsymbol{E}^{\mathrm{inc},(m)}$ in the training
set, the network produces a predicted surface current
$\boldsymbol{J}_{\mathrm{pred}}^{(m)}$ through
\eqref{eq:Jpred_assemble}, with
$\boldsymbol{J}_{\mathrm{near}}^{(m)}$ and
$\boldsymbol{J}_{\mathrm{far}}^{(m)}$ produced by the
near- and far-field branches of
Sections~\ref{subsec:nearfield_func}
and~\ref{subsec:farfield_func}. The corresponding
scattered field at each collocation
point~$\boldsymbol{p}_k\in\mathcal{P}$ is evaluated by
acting the discretized EFIE operator on all source
points~$\boldsymbol{p}_j\in\mathcal{P}$, and the resulting
per-excitation residual reads as follows:
\begin{equation}
\begin{split}
\mathcal{L}^{(m)} = \frac{1}{K} \sum_{k=1}^{K}&\left\|\hat{\boldsymbol{n}}(\boldsymbol{p}_k)\right.\\ &\left.\times \Big( \boldsymbol{E}^{(m)}_{\text{inc}}(\boldsymbol{p}_k) + \sum_{j=1}^{K} \mathbf{G}(\boldsymbol{p}_k,\boldsymbol{p}_j) \boldsymbol{J}^{(m)}_{\mathrm{pred}}(\boldsymbol{p}_j) \Big) \right\|_2^2 ,
\label{eq:esca_discrete}
\end{split}
\end{equation}
where $K=|\mathcal{P}|$, and
$\hat{\boldsymbol{n}}(\boldsymbol{p}_k)$ is the outward
unit normal at $\boldsymbol{p}_k$. With a slight abuse of
notation, the kernel
$\mathbf{G}(\boldsymbol{p}_k,\boldsymbol{p}_j)$ in the
inner sum denotes the quadrature-weighted discrete EFIE
operator from the source point~$\boldsymbol{p}_j$ to the
observation point~$\boldsymbol{p}_k$: for regular
interactions it reduces to the product of the dyadic
kernel with the surface area element~$\Delta s_j$, whereas
the self and near-singular terms are handled by standard
singularity-extraction techniques following MoM
practice~\cite{greenfunction}.

The inner summation
therefore represents the scattered field produced by
$\boldsymbol{J}_{\mathrm{pred}}^{(m)}$ at
$\boldsymbol{p}_k$, and $\mathcal{L}^{(m)}$ measures the
extent to which the tangential total field departs from
the PEC boundary condition over the collocation set. The summation over $j=1,\ldots,K$ in~\eqref{eq:esca_discrete}
runs over the full surface point set $\mathcal{P}$, and should not
be confused with the local neighborhood summation in the
near-field message passing of~\eqref{eq:gnn_update}:
the former evaluates the discretized EFIE operator for residual
computation, whereas the latter realizes the short-range
learnable interaction in the near-field branch.

The overall physics-informed loss is obtained by averaging
the per-excitation residual over a batch of~$N_s$ incident
fields,
\begin{equation}
    \mathcal{L}_{\mathrm{phys}}
    =
    \frac{1}{N_s}
    \sum_{m=1}^{N_s}
    \mathcal{L}^{(m)},
    \label{eq:physics_loss}
\end{equation}
and is used as the sole training objective for U-PINet.
The connection to the classical solver is conceptual: a
conventional MLFMA solver iteratively reduces the discrete
residual
$\boldsymbol{r}=\boldsymbol{V}-\mathbf{Z}\boldsymbol{I}$
of a fixed linear system for a single scattering
instance, whereas U-PINet minimizes
$\mathcal{L}_{\mathrm{phys}}$ over the network parameters
across a training set of incident fields, so that after
training a single forward pass produces a current
distribution that approximately satisfies the discretized
EFIE on previously unseen but related configurations.
Once~$\boldsymbol{J}_{\mathrm{pred}}$ has been obtained, it
is substituted into the radiation
integral~\eqref{eq:Esca_operator} to evaluate the
scattered field at exterior observation points, from which
the bistatic RCS~$\sigma(\theta,\varphi)$ is computed
via~\eqref{eq:rcs_def}.

\section{The U-PINet Neural Network Architecture}
\label{sec:architecture}

This section presents the explicit network operations that
realize the functional operators
$\mathcal{K}(\cdot)$, $\Phi^{(t)}$, $\boldsymbol{\Gamma}_k$,
$\boldsymbol{\mathcal{A}}_k$, and~$\boldsymbol{\Psi}_k$
introduced in Section~\ref{sec:UPINET_framework}. As shown
in Fig.~\ref{fig:test2}, the network takes the surface
point cloud, the surface normals, and the incident field
as inputs, and outputs the predicted surface
current~$\boldsymbol{J}_{\mathrm{pred}}$ in a single
forward pass. The implementation is organized into a
near-field graph encoder
(Section~\ref{subsec:nf_encoder}) and a hierarchical
far-field fusion module
(Section~\ref{sec:far_field_module}), followed by a brief
description of the resulting end-to-end forward pass
(Section~\ref{subsec:forward_diff}).

\begin{figure}[!t]
\centering
\includegraphics[width=8.5cm]{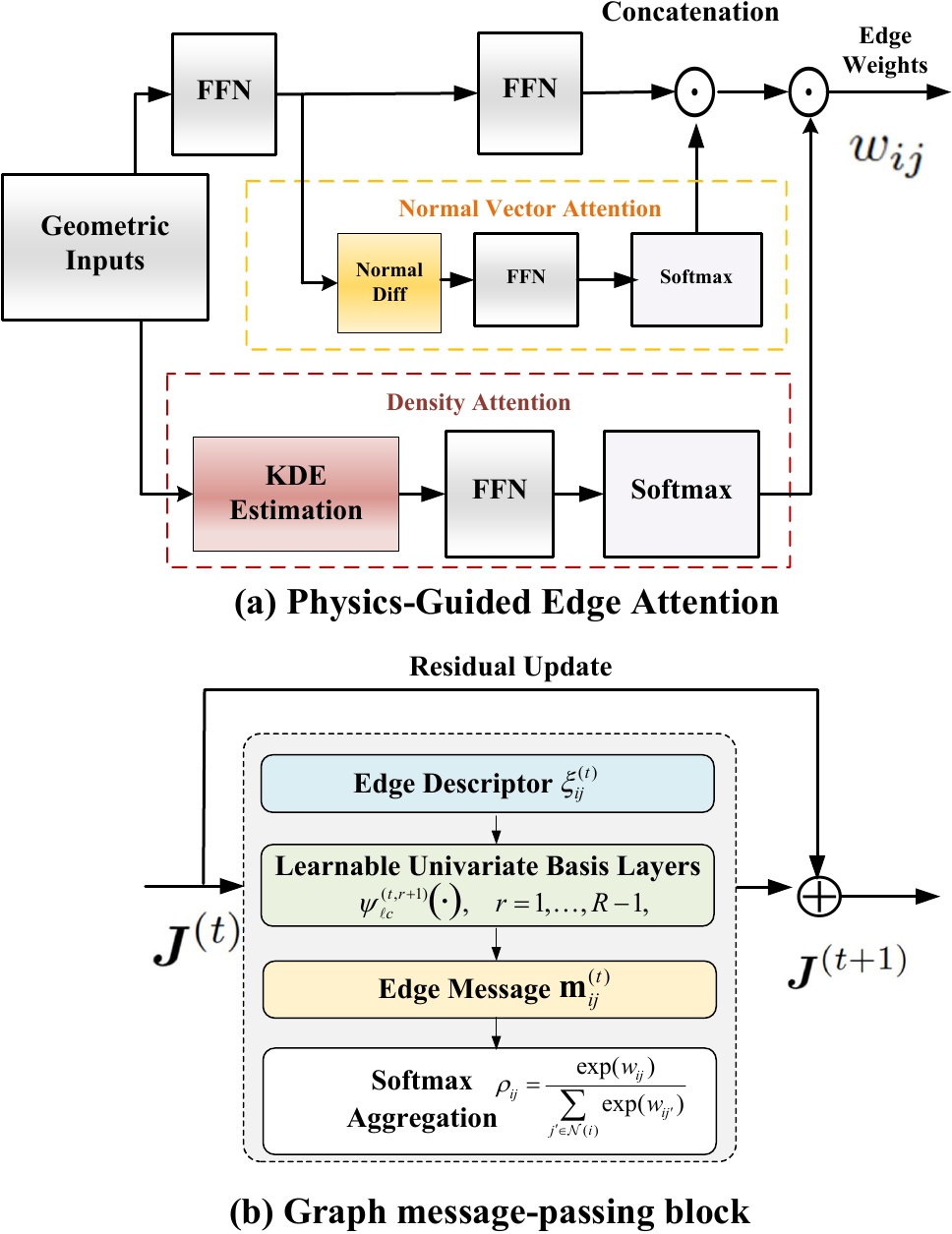}
\caption{Internal structure of the near-field graph
encoder module. (a) Physics-guided edge attention
combining a kernel-density-based branch (producing
$\beta_{ij}$) and a surface-normal-based branch
(producing~$\gamma_{ij}$) into the scalar edge
weight~$w_{ij}$, realizing
$\mathcal{K}(\cdot)$ in~\eqref{eq:edge_feat}.
(b) Graph message-passing block in which the edge
descriptor~$\boldsymbol{\xi}_{ij}^{(t)}$ is processed by
learnable univariate basis layers to generate the edge
message~$\boldsymbol{m}_{ij}^{(t)}$, followed by softmax
aggregation with weights~$\rho_{ij}$ and a residual node
update, realizing~$\Phi^{(t)}$
in~\eqref{eq:edge_message}.}
\label{fig: test3}
\end{figure}

\subsection{Near-Field Graph Encoder}
\label{subsec:nf_encoder}

This module realizes the edge
kernel~$\mathcal{K}(\cdot)$ of~\eqref{eq:edge_feat} and
the message kernel~$\Phi^{(t)}$
of~\eqref{eq:edge_message} through two cooperating blocks.

\subsubsection{Physics-Guided Edge Attention}

The edge kernel~$\mathcal{K}(\cdot)$ is implemented as a
dual-branch attention module that combines local sampling
density and surface-normal differences in Fig.~\ref{fig: test3}(a). The local density at each
point~$\boldsymbol{p}_i$ is estimated by Gaussian kernel
density estimation~\cite{ref35}
over~$\mathcal{N}(i)$:
\begin{equation}
\widehat{f}(\bm{p}_i) =
\frac{1}{N_i (2\pi)^{3/2} h^3}
\sum_{j\in\mathcal{N}(i)}
\exp\left(
-\frac{\|\bm{p}_i-\bm{p}_j\|^2}{2h^2}
\right),
\label{eq:kde}
\end{equation}
with $N_i=|\mathcal{N}(i)|$ and bandwidth~$h>0$. The
normalized inverse density
$d_j=\mathcal{D}(\boldsymbol{p}_j)/\max_{\ell}\mathcal{D}(\boldsymbol{p}_{\ell})$,
where
$\mathcal{D}(\boldsymbol{p}_i)=1/\widehat{f}(\boldsymbol{p}_i)$,
is mapped through a two-layer feed-forward network~$F_d$
to a density-attention coefficient
\begin{equation}
    \beta_{ij}
    =
    \frac{\exp\!\big(F_d(d_j)\big)}
    {\sum_{j'\in\mathcal{N}(i)}\exp\!\big(F_d(d_{j'})\big)}.
    \label{eq:density_attn}
\end{equation}
In parallel, the surface-normal
difference~$\Delta\hat{\boldsymbol{n}}_{ij}$ is mapped
through a second feed-forward network~$F_n$ to a
normal-attention coefficient
\begin{equation}
    \gamma_{ij}
    =
    \frac{\exp\!\big(F_n(\Delta\hat{\boldsymbol{n}}_{ij})\big)}
    {\sum_{j'\in\mathcal{N}(i)}
    \exp\!\big(F_n(\Delta\hat{\boldsymbol{n}}_{ij'})\big)}.
    \label{eq:normal_attn}
\end{equation}
The density coefficient~$\beta_{ij}$ reflects the local
sampling regularity, while the normal
coefficient~$\gamma_{ij}$ reflects the geometric
anisotropy across the edge; both quantities are known to
correlate with the magnitude of pairwise impedance entries
in MoM. The two coefficients are concatenated and
projected through a multilayer perceptron to obtain the
scalar edge weight
\begin{equation}
    w_{ij} = \mathrm{MLP}(\beta_{ij}\oplus\gamma_{ij}),
    \label{eq:edge_weight_impl}
\end{equation}
which enters the edge
descriptor~\eqref{eq:edge_descriptor} and modulates the
contribution of neighbor~$j$ in the subsequent
message-passing step.
\subsubsection{Iterative Message-Passing Update with Learnable Univariate Basis Functions}
The message kernel $\Phi^{(t)}$ is realized through a
basis-function block parameterized by learnable univariate
nonlinearities. Motivated by the Kolmogorov--Arnold
representation theorem~\cite{kan_theorem},
which expresses any continuous multivariate function as a
finite superposition of univariate functions, we adopt the
learnable basis-function parameterization
\begin{equation}
    \Phi^{(t)}\bigl(\boldsymbol{\xi}_{ij}^{(t)}\bigr)
    = \sum_{q=1}^{Q} a_q^{(t)}\,\phi_q\!\bigl(
        (\boldsymbol{u}_q^{(t)})^{\top}\boldsymbol{\xi}_{ij}^{(t)} + b_q^{(t)}
      \bigr),
    \label{eq:phi_shallow_KA}
\end{equation}
where $\{\phi_q\}_{q=1}^{Q}$ are univariate
nonlinearities and
$\{a_q^{(t)},\boldsymbol{u}_q^{(t)},b_q^{(t)}\}$ are
the associated parameters. This parameterization places the learnable degrees of freedom on the univariate activations rather than on fixed nonlinearities, equipping the message kernel with a geometry-adaptive nonlinear shape that is well suited to representing the spatially varying near-field coupling on a discretized PEC surface. To enhance the expressive
capacity required by the multi-feature edge descriptor
in~\eqref{eq:edge_descriptor}, we adopt the deep
extension of~\eqref{eq:phi_shallow_KA}~\cite{ref41}, in which
the univariate nonlinearities are parameterized as
learnable splines~$\psi_{\ell c}^{(t,r)}(\cdot)$ and
stacked across~$R$ layers in Fig.~\ref{fig: test3}(b):
\begin{equation}
\begin{aligned}
    z_{ij,c}^{(1)}
    &=
    \sum_{\ell=1}^{d_{\xi}}
    \psi_{\ell c}^{(t,1)}\!\big(\xi_{ij,\ell}^{(t)}\big),
    \quad c=1,\ldots,d_1, \\
    z_{ij,c}^{(r+1)}
    &=
    \sum_{\ell=1}^{d_r}
    \psi_{\ell c}^{(t,r+1)}\!\big(z_{ij,\ell}^{(r)}\big),
    \quad r=1,\ldots,R-1, \\
    \boldsymbol{m}_{ij}^{(t)}
    &=
    \mathbf{W}_{\mathrm{o}}^{(t)}\,\boldsymbol{z}_{ij}^{(R)},
\end{aligned}
\label{eq:gkan_edge_message}
\end{equation}
where $\psi_{\ell c}^{(t,r)}(\cdot)$ is a learnable
univariate spline at the $r$-th layer of iteration~$t$,
and~$\mathbf{W}_{\mathrm{o}}^{(t)}$ is an output
projection that matches the dimensions of the node feature
space. The output~$\boldsymbol{m}_{ij}^{(t)}$
of~\eqref{eq:gkan_edge_message} is the edge message
in~\eqref{eq:edge_message}, completing the realization of
$\Phi^{(t)}$. The shallow form~\eqref{eq:phi_shallow_KA}
serves as the theoretical anchor and is recovered as the
$R=1$ special case of~\eqref{eq:gkan_edge_message}, in
which a single layer of univariate activations is followed
by the linear readout.

The edge messages are then aggregated at
node~$\boldsymbol{p}_i$ using the edge
weights~$\{w_{ij}\}$ from~\eqref{eq:edge_weight_impl},
with a softmax normalization for numerical stability:
\begin{equation}
    \Delta\boldsymbol{J}^{(t)}(\boldsymbol{p}_i)
    =
    \sum_{j\in\mathcal{N}(i)}
        \rho_{ij}\,\boldsymbol{m}_{ij}^{(t)},
    \quad
    \rho_{ij}
    =
    \frac{\exp(w_{ij})}
    {\sum_{j'\in\mathcal{N}(i)}\exp(w_{ij'})}.
    \label{eq:gkan_aggregation}
\end{equation}
The current at iteration~$t+1$ is obtained by the residual
update
$\boldsymbol{J}^{(t+1)}(\boldsymbol{p}_i)
= \boldsymbol{J}^{(t)}(\boldsymbol{p}_i)
+ \Delta\boldsymbol{J}^{(t)}(\boldsymbol{p}_i)$, which is
the explicit form of the abstract
update~\eqref{eq:gnn_update} once the
weight~$\rho_{ij}$ from the physics-guided edge attention
is incorporated. After~$T_{\mathrm{iter}}$
iterations, the near-field output is read out as
\begin{equation}
    \boldsymbol{J}_{\mathrm{near}}
    = \boldsymbol{J}^{(T_{\mathrm{iter}})},
    \label{eq:J_near_gkan}
\end{equation}
and is forwarded to the far-field fusion module as the
input at the finest hierarchical level.

\subsection{Hierarchical Far-Field Fusion Module}
\label{sec:far_field_module}

This module realizes the multilevel
operators~$\boldsymbol{\Gamma}_k$, $\boldsymbol{\mathcal{A}}_k$,
and~$\boldsymbol{\Psi}_k$ defined
in~\eqref{eq:upinet_P}, \eqref{eq:upinet_Phi_def},
and~\eqref{eq:upinet_disagg}. All three operators share
the same octree
partition~$\{\mathcal{C}^{(k)}\}_{k=1}^{L}$ introduced in
Section~\ref{subsec:farfield_func}, which provides a
hierarchical organization of the surface unknowns from
the finest level~$k=L$ to the coarsest level~$k=1$.

\subsubsection{Pooling}

Starting from
$\boldsymbol{S}^{(L)}=\boldsymbol{J}_{\mathrm{near}}$, the
descriptor of each octree cell~$\mathcal{C}^{(k)}_c$ at
level~$k=L-1,\ldots,1$ is obtained by combining the mean
and the maximum of its child features:
\begin{equation}
    \boldsymbol{S}^{(k)}_c
    =
    \mathbf{W}_{\Gamma}^{(k)}
    \,\mathrm{cat}\!\left[
        \tfrac{1}{|\mathcal{C}^{(k)}_c|}\!\!
        \sum_{i\in\mathcal{C}^{(k)}_c}\!\!\boldsymbol{S}^{(k+1)}_i,\;
        \max_{\,i\in\mathcal{C}^{(k)}_c}\boldsymbol{S}^{(k+1)}_i
    \right],
    \label{eq:pooling}
\end{equation}
where $\mathrm{cat}[\cdot,\cdot]$ stacks the two vectors
along the feature axis and~$\mathbf{W}_{\Gamma}^{(k)}$ is
a trainable linear projection that produces a compact
summary of each parent cell. The mean term preserves the
average behavior of the child cells, while the maximum
term retains the most salient response within each cell;
their concatenation gives a richer descriptor than either
alone for representing the coarser-level source.

\subsubsection{Multi-Head Cross-Attention}
\label{subsubsec:phik_attn}

The cross-scale coupling
operator~$\boldsymbol{\mathcal{A}}_k$
in~\eqref{eq:upinet_Phi_def} is realized as a multi-head
attention block. At each head, the level-$k$
features~$\boldsymbol{S}^{(k)}$ are used as queries, while
the unpooled coarser
context~$\mathcal{U}_k(\tilde{\boldsymbol{S}}^{(k-1)})$
supplies both keys and values:
\begin{align}
    \boldsymbol{Q}^{(k)}
    &= \mathbf{W}_Q^{(k)}\,\boldsymbol{S}^{(k)},
    \label{eq:Q}\\
    \boldsymbol{K}^{(k)}
    &= \mathbf{W}_K^{(k)}\,
       \mathcal{U}_k(\tilde{\boldsymbol{S}}^{(k-1)}),
    \label{eq:K}\\
    \boldsymbol{V}^{(k)}
    &= \mathbf{W}_V^{(k)}\,
       \mathcal{U}_k(\tilde{\boldsymbol{S}}^{(k-1)}),
    \label{eq:V}
\end{align}
through trainable projection
matrices~$\mathbf{W}_Q^{(k)}$, $\mathbf{W}_K^{(k)}$,
and~$\mathbf{W}_V^{(k)}$. The single-head attention
weights and the corresponding head output are then
\begin{equation}
    a^{(k)}_{i,m}
    =
    \frac{
        \exp\!\big(\langle \boldsymbol{Q}^{(k)}_i,\,
        \boldsymbol{K}^{(k)}_m\rangle / \sqrt{d_k}\big)
    }{
        \sum_{m'} \exp\!\big(\langle
        \boldsymbol{Q}^{(k)}_i,\,
        \boldsymbol{K}^{(k)}_{m'}\rangle / \sqrt{d_k}\big)
    },
    \quad
    \boldsymbol{O}^{(k)}_i
    =
    \sum_m a^{(k)}_{i,m}\,\boldsymbol{V}^{(k)}_m,
    \label{eq:att_combined}
\end{equation}
where $d_k$ is the per-head feature dimension and the
data-dependent weights~$a^{(k)}_{i,m}$ allow each
fine-scale element to selectively integrate information
from coarser-scale features, which is the mechanism by
which $\boldsymbol{\mathcal{A}}_k$ captures long-range coupling
between distant parts of the surface. With $H$ parallel
heads of identical structure but independent projection
matrices, the head outputs are concatenated and combined
with the level-$k$ input through a residual connection,
\begin{align}
    \boldsymbol{H}^{(k)}_{\mathrm{attn}}
    &= \boldsymbol{S}^{(k)}
       + \mathrm{MHA}_H\!\Big(
            \boldsymbol{S}^{(k)},\,
            \mathcal{U}_k(\tilde{\boldsymbol{S}}^{(k-1)}),\,
            \mathcal{U}_k(\tilde{\boldsymbol{S}}^{(k-1)})
         \Big),
    \label{eq:far_attn_layer}\\
    \tilde{\boldsymbol{S}}^{(k)}
    &= \boldsymbol{H}^{(k)}_{\mathrm{attn}}
       + \mathrm{FFN}\!\big(\mathrm{LN}(\boldsymbol{H}^{(k)}_{\mathrm{attn}})\big),
    \label{eq:far_attn_ffn}
\end{align}
where~$\mathrm{MHA}_H(\cdot,\cdot,\cdot)$ denotes the
$H$-head version of the operation
in~\eqref{eq:Q}--\eqref{eq:att_combined} with its three
arguments taken as the query, key, and value sequences,
$\mathrm{LN}(\cdot)$ is layer normalization,
and~$\mathrm{FFN}(\cdot)$ is a two-layer feed-forward
network. The unpooling operator~$\mathcal{U}_k$ assigns
each parent-cell descriptor to all of its child cells
in~$\mathcal{C}^{(k)}$, so that fine-level features at
level~$k$ can interact with the coarser context lifted
from level~$k-1$.

\subsubsection{Kernel-Point Fusion}
\label{subsubsec:psik_kpconv}

The fusion operator~$\boldsymbol{\Psi}_k$
in~\eqref{eq:upinet_disagg} first lifts the coarser-level
far-field current~$\boldsymbol{J}^{(k-1)}_{\mathrm{far}}$
to level~$k$,
\begin{equation}
    \hat{\boldsymbol{J}}^{(k-1\to k)}_{\mathrm{far}}(\boldsymbol{p}_i)
    = \mathcal{U}_k\!\big(\boldsymbol{J}^{(k-1)}_{\mathrm{far}}\big)(\boldsymbol{p}_i),
    \quad \boldsymbol{p}_i \in \mathcal{C}^{(k)},
    \label{eq:disagg_unpool}
\end{equation}
and then introduces local spatial dependence within the
unpooled neighborhood through a kernel-point
mapping~\cite{thomas2019kpconv}. With~$N_{\mathrm{kern}}$
fixed kernel anchors~$\{\boldsymbol{\zeta}_q\}$ placed in
the local frame of~$\boldsymbol{p}_i$, the level-$k$
far-field current is updated as
\begin{equation}
    \boldsymbol{J}^{(k)}_{\mathrm{far}}(\boldsymbol{p}_i)
    =
    \tilde{\boldsymbol{S}}^{(k)}_i
    +
    \sum_{q=1}^{N_{\mathrm{kern}}}
        \rho_q(\boldsymbol{p}_i)\,
        \mathbf{W}^{(k)}_{q}\,
        \hat{\boldsymbol{J}}^{(k-1\to k)}_{\mathrm{far}}(\boldsymbol{p}_i),
    \label{eq:far_disagg_impl}
\end{equation}
where $\mathbf{W}^{(k)}_{q}$ are learnable kernel
matrices and
\begin{equation}
    \rho_q(\bm{p}_i) = \max\bigl(0,\; 1 - \|\bm{p}_i - \bm{\zeta}_q\|/\sigma_\kappa\bigr),
    \label{eq:kpconv_kernel}
\end{equation}
is a linear correlation kernel with bandwidth~$\sigma_\kappa$.
The kernel-point construction injects a position-dependent
weighting into the disaggregation step, so that distinct
locations within the same parent cell receive
distinguishable updates rather than a uniform copy of the
coarser-level current. Cascading~\eqref{eq:pooling}
through~\eqref{eq:far_disagg_impl} from $k=2$ up to $k=L$
yields the finest-level far-field
current~$\boldsymbol{J}_{\mathrm{far}}
=\boldsymbol{J}^{(L)}_{\mathrm{far}}$.

\subsection{Training Objective and Optimization}
\label{subsec:forward_diff}

The components introduced in
Sections~\ref{subsec:nf_encoder} and~\ref{sec:far_field_module}
together define a single forward pass of U-PINet, which
takes the surface point cloud~$\mathcal{P}$, the
associated outward normals~$\{\hat{\boldsymbol{n}}_i\}$,
and a sampled incident field~$\boldsymbol{E}^{\mathrm{inc}}$
as inputs and returns the predicted surface
current~$\boldsymbol{J}_{\mathrm{pred}}$
in~\eqref{eq:Jpred_assemble} as the output. The near-field
graph encoder yields~$\boldsymbol{J}_{\mathrm{near}}$
in~\eqref{eq:J_near_gkan}, while the hierarchical
far-field fusion module
yields~$\boldsymbol{J}_{\mathrm{far}}$ through the cascade
of~\eqref{eq:pooling}--\eqref{eq:far_disagg_impl}.

A direct evaluation of~\eqref{eq:esca_discrete} has a nominal
complexity of $\mathcal{O}(K^{2})$ per excitation, since each
of the $K$ collocation points sums the contributions of all
$K$ source points through the discretized EFIE operator. This
cost becomes prohibitive on the electrically larger targets
considered in Section~\ref{section:E}. To keep the
training tractable, at each optimization step we sample a
random subset $\mathcal{P}_c \subset \mathcal{P}$ of
$K_c \ll K$ collocation points and evaluate the residual only
on $\mathcal{P}_c$, while the inner source summation in
\eqref{eq:esca_discrete} is retained over the full point
set~$\mathcal{P}$. This reduces the per-step cost from
$\mathcal{O}(K^{2})$ to $\mathcal{O}(K_c K)$ and provides an
unbiased estimator of the full residual in expectation. The
subset $\mathcal{P}_c$ is resampled at every step so that all
collocation points contribute to the parameter update over a
training epoch.

\begin{figure}[!t]
\centering
\includegraphics[scale=0.25]{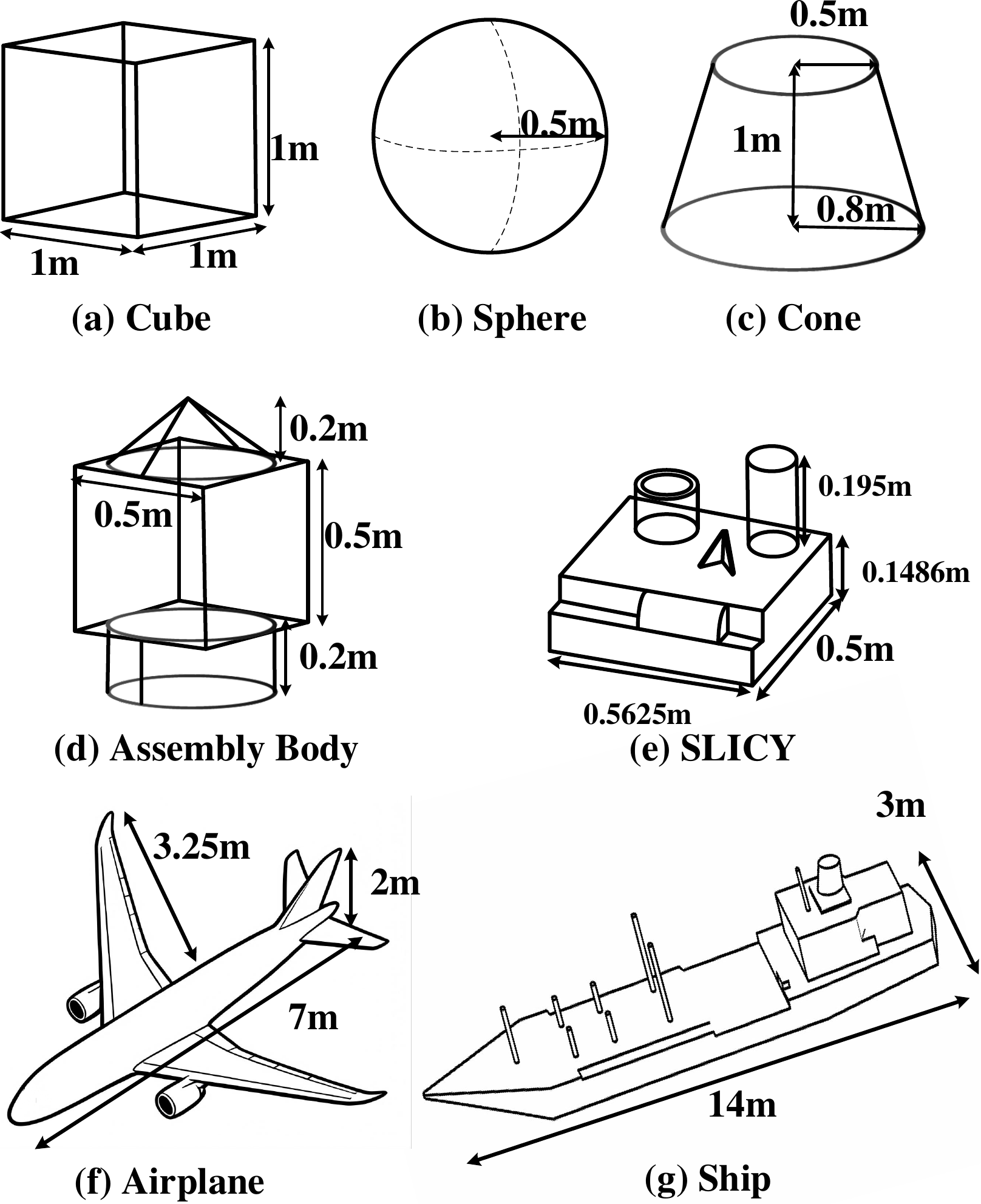}
\caption{Canonical and complex 3D PEC targets considered for
the evaluation of U-PINet, including cube, sphere, cone,
assembly body, SLICY, airplane, and ship.}
\label{fig: basic}
\end{figure}

\section{Experimental Setup}
\label{section:D}

This section describes the experimental setup used to
evaluate the accuracy and physical consistency of U-PINet.
As shown in Fig.~\ref{fig: basic}, seven 3D PEC targets
are considered: four canonical targets (cube, sphere,
cone, and assembly body) and three complex targets (SLICY
\cite{slicy}, airplane, and ship). Together, these
geometries cover smooth surfaces, sharp edges,
multi-component structures, and electrically larger
targets, providing a representative testbed for surface
current reconstruction and bistatic RCS prediction at
microwave frequencies.

\subsection{Data Generation}
\label{subsec:data_gen}

For each target, the triangular surface mesh is converted
into a point-cloud representation in which the centroid of
each triangular element is taken as a surface point. The network input at each point consists of its 3D coordinates,
the outward surface normal, the local geometric descriptors
used in the edge-attention block (the kernel-density-based
local sampling regularity and the pairwise surface-normal
difference $\Delta\hat{\boldsymbol{n}}_{ij}$, as defined in
Section~\ref{subsec:nearfield_func}), and the incident
plane-wave vector $\boldsymbol{E}^{\mathrm{inc}}(\boldsymbol{p})$
sampled at that point. The network predicts the induced surface
current on the PEC surface, from which the far-field RCS
is computed using the radiation
integral~\eqref{eq:Esca_operator}
and~\eqref{eq:rcs_def}.

The targets are illuminated by a plane wave of amplitude
$1$~V/m. Unless otherwise stated, the operating frequency
is $1$~GHz. The airplane target is additionally evaluated
at three frequency points ($0.5$, $1$, and $2$~GHz) to
examine multi-frequency behavior. HH and VV co-polarized
configurations are considered for selected targets. The
incident and observation angles follow the same sampling
grid for all compared methods, and MLFMA is used to
generate reference surface currents and RCS responses for
evaluation and visualization purposes only; these
reference currents do not enter the training process as
labels.

Throughout the remainder of this paper, MLFMA solutions are
adopted as the numerical ground truth for quantitative evaluation
and visualization. It should be emphasized that the proposed U-PINet
is trained in a physics-driven, label-free manner through the EFIE
residual in~\eqref{eq:physics_loss}, without access to MLFMA-generated currents or
RCS values; the MLFMA data are used solely as an independent
benchmark for assessing the trained model.

The average edge length of the triangular mesh is set
to approximately $\lambda_{\mathrm{ref}}/10$ for the canonical
and moderately complex targets, where
$\lambda_{\mathrm{ref}} = c_0/f$ denotes the free-space
wavelength at the operating frequency $f$ of the corresponding
experiment, and $c_0$ is the speed of light in vacuum.
For the electrically larger airplane and ship targets, a
coarser discretization is adopted in order to keep the
number of unknowns tractable.

\subsection{Benchmark Models and Evaluation Metrics}
\label{subsec:metrics}

U-PINet is compared with two representative
physics-informed CEM baselines, denoted MLFMA-PINN and
PhiGRL. To ensure a controlled comparison, all models
share the same point-cloud input representation,
incident-field encoding, training/test split, and
evaluation protocol; baseline-specific design choices that
differ from the original references are noted below.

\begin{itemize}
\item \textbf{MLFMA-PINN}~\cite{ref22}: This baseline
represents the supervised regression paradigm, in which
neural modules are trained to reproduce MLFMA reference
solutions on a per-target basis. Following the original
design, a generalized regression neural network (GRNN)
branch with a Gaussian kernel handles the
accuracy-sensitive translation components, and an ANN
branch processes the remaining components through a stack
of fully connected layers with tangent-sigmoidal hidden
activations and a linear output layer; the input
translation variables are normalized to $[-1,1]$. The
training objective is the mean squared error between the
predicted and the MLFMA reference surface currents on the
training incident-angle configurations. This baseline is
included as a representative of the label-supervised
learning paradigm, against which the proposed
label-free training of U-PINet can be assessed.

\item \textbf{PhiGRL}~\cite{ref28}: This baseline follows
the physics-informed graph residual learning framework
proposed for 3D PEC scattering. The surface mesh is
represented as a graph whose nodes are the surface
unknowns and whose edges are induced by local mesh
connectivity. Following the original architecture, two
parallel graph neural network branches predict the real
and imaginary parts of the induced current; each branch
consists of three graph convolutional layers with mean
aggregation, graph batch normalization, and ELU
activation, followed by dropout and a linear output
layer. The network is trained against the
physics-informed loss defined in the original work,
which measures the residual of the discretized MoM
system associated with the predicted current
coefficients. No MLFMA reference currents are used as
supervision.

\begin{table*}[!t]
\centering
\caption{Overall Accuracy Comparison of Surface Current
Reconstruction and RCS Prediction on Canonical and
Complex 3D PEC Targets at $1$~GHz, Averaged Over HH and
VV Polarizations and Over the Test Incident and
Observation Angles.}
\label{tab:overall_accuracy}
\renewcommand{\arraystretch}{1.10}
\setlength{\tabcolsep}{2.5pt}
\footnotesize
\begin{adjustbox}{max width=\textwidth}
\begin{tabular}{lcccccccccccc}
\toprule
\multirow{2}{*}{Target}
& \multicolumn{4}{c}{MLFMA-PINN}
& \multicolumn{4}{c}{PhiGRL}
& \multicolumn{4}{c}{U-PINet (Proposed)} \\
\cmidrule(lr){2-5}
\cmidrule(lr){6-9}
\cmidrule(lr){10-13}
& $\mathrm{NMSE}_{J}\downarrow$
& $\mathrm{MAE}_{J}$~(A/m)$\downarrow$
& $\mathrm{RMSE}_{\mathrm{RCS}}$~(dB)$\downarrow$
& $R^2_{\mathrm{RCS}}\uparrow$
& $\mathrm{NMSE}_{J}\downarrow$
& $\mathrm{MAE}_{J}$~(A/m)$\downarrow$
& $\mathrm{RMSE}_{\mathrm{RCS}}$~(dB)$\downarrow$
& $R^2_{\mathrm{RCS}}\uparrow$
& $\mathrm{NMSE}_{J}\downarrow$
& $\mathrm{MAE}_{J}$~(A/m)$\downarrow$
& $\mathrm{RMSE}_{\mathrm{RCS}}$~(dB)$\downarrow$
& $R^2_{\mathrm{RCS}}\uparrow$ \\
\midrule
Cube
& $2.14 \times 10^{-3}$ & 0.0242 & 0.88 & 0.951
& $1.08 \times 10^{-3}$ & 0.0178 & 0.79 & 0.974
& $4.80 \times 10^{-4}$ & 0.0125 & 0.68 & 0.981 \\
Sphere
& $1.46 \times 10^{-3}$ & 0.0226 & 1.05 & 0.884
& $9.70 \times 10^{-4}$ & 0.0165 & 0.88 & 0.941
& $4.50 \times 10^{-4}$ & 0.0118 & 0.77 & 0.972 \\
Cone
& $2.45 \times 10^{-3}$ & 0.0282 & 0.81 & 0.942
& $1.25 \times 10^{-3}$ & 0.0204 & 0.72 & 0.968
& $5.90 \times 10^{-4}$ & 0.0150 & 0.65 & 0.985 \\
Assembly Body
& $6.12 \times 10^{-3}$ & 0.0584 & 1.18 & 0.865
& $3.64 \times 10^{-3}$ & 0.0442 & 0.98 & 0.928
& $1.85 \times 10^{-3}$ & 0.0345 & 0.84 & 0.971 \\
SLICY
& $8.42 \times 10^{-3}$ & 0.0768 & 1.04 & 0.912
& $5.76 \times 10^{-3}$ & 0.0582 & 0.96 & 0.948
& $3.15 \times 10^{-3}$ & 0.0482 & 0.92 & 0.965 \\
Airplane
& $1.28 \times 10^{-2}$ & 0.0984 & 2.64 & 0.804
& $8.42 \times 10^{-3}$ & 0.0782 & 2.31 & 0.871
& $5.00 \times 10^{-3}$ & 0.0590 & 2.15 & 0.915 \\
Ship
& $1.62 \times 10^{-2}$ & 0.1246 & 3.62 & 0.712
& $1.05 \times 10^{-2}$ & 0.0924 & 3.21 & 0.814
& $5.65 \times 10^{-3}$ & 0.0671 & 2.95 & 0.865 \\
\midrule
Average
& $7.08 \times 10^{-3}$ & 0.0619 & 1.60 & 0.867
& $4.52 \times 10^{-3}$ & 0.0468 & 1.41 & 0.921
& $2.45 \times 10^{-3}$ & 0.0354 & 1.28 & 0.951 \\
\bottomrule
\end{tabular}
\end{adjustbox}
\vspace{0.5mm}
\footnotesize{
Note: $\mathrm{NMSE}_{J}$ is dimensionless;
$\mathrm{MAE}_{J}$ is reported in A/m;
$\mathrm{RMSE}_{\mathrm{RCS}}$ is reported in the
dB-domain root-mean-square error of $\sigma$;
$R^2_{\mathrm{RCS}}$ is dimensionless.
$\downarrow$/$\uparrow$ indicate that lower/higher values
are better.
}
\end{table*}

\item \textbf{U-PINet (proposed)}: The proposed model
combines the near-field graph encoder of
Section~\ref{subsec:nf_encoder} and the hierarchical
far-field fusion module of
Section~\ref{sec:far_field_module}. The near-field
neighborhood radius is set to $r_{\mathrm{nf}} =
0.5\lambda_{\mathrm{ref}}$, with hidden feature
dimension $128$ and $T_{\mathrm{iter}} = 3$
message-passing iterations. The far-field branch uses an
octree of $L = 4$ levels, $H = 4$ attention heads, and
hidden dimension $256$. Training follows the
physics-informed objective defined
in~\eqref{eq:physics_loss}.
\end{itemize}

Reconstruction accuracy is assessed at both the current
and RCS levels. For the induced surface current, the
normalized mean square error (NMSE) and mean absolute
error (MAE) with respect to the MLFMA reference
$\boldsymbol{J}_{\mathrm{ref}}$ are defined as
\begin{equation}
    \mathrm{NMSE}_{\boldsymbol{J}}
    =
    \frac{\sum_{i}\|\boldsymbol{J}_{\mathrm{pred}}(\boldsymbol{p}_i)
        - \boldsymbol{J}_{\mathrm{ref}}(\boldsymbol{p}_i)\|^{2}}
         {\sum_{i}\|\boldsymbol{J}_{\mathrm{ref}}(\boldsymbol{p}_i)\|^{2}},
\end{equation}
\begin{equation}
    \mathrm{MAE}_{\bm{J}} = \frac{1}{K}\sum_{i=1}^{K}\bigl\|\bm{J}_{\mathrm{pred}}(\bm{p}_i)-\bm{J}_{\mathrm{ref}}(\bm{p}_i)\bigr\|.
\end{equation}
For the RCS response, the root-mean-square error in the
$\mathrm{dB}$ domain and the coefficient of determination
are reported,
\begin{equation}
    \mathrm{RMSE}_{\mathrm{RCS}}
    =
    \sqrt{
        \frac{1}{M}\sum_{m=1}^{M}
        \big(\sigma_{\mathrm{pred}}^{\mathrm{dB}}(\theta_m,\varphi_m)
        - \sigma_{\mathrm{ref}}^{\mathrm{dB}}(\theta_m,\varphi_m)\big)^{2}
    },
\end{equation}
\begin{equation}
    R^{2}_{\mathrm{RCS}}
    =
    1
    -
    \frac{\sum_{m}\big(\sigma_{\mathrm{pred}}^{\mathrm{dB}}(\theta_m,\varphi_m)
        - \sigma_{\mathrm{ref}}^{\mathrm{dB}}(\theta_m,\varphi_m)\big)^{2}}
         {\sum_{m}\big(\sigma_{\mathrm{ref}}^{\mathrm{dB}}(\theta_m,\varphi_m)
        - \bar{\sigma}_{\mathrm{ref}}^{\mathrm{dB}}\big)^{2}},
\end{equation}
where $M$ is the number of evaluated observation
directions and $\bar{\sigma}_{\mathrm{ref}}^{\mathrm{dB}}$
denotes the mean of the reference values over the same
set. Unless otherwise stated, all metrics reported in
Section~\ref{section:E} are computed under matched
geometry, frequency, polarization, incident-angle, and
observation-angle settings.

\subsection{Training Configuration}
\label{subsec:training_config}

All learning-based models are implemented in PyTorch and
trained on a single NVIDIA RTX 3090Ti GPU. For each
target, the available samples are partitioned into
training and test sets with an $80{:}20$ ratio, with the
test set disjoint from the training set in incident-angle
configurations. Each model is trained against the
objective associated with its original formulation:
MLFMA-PINN minimizes the mean squared error with respect
to MLFMA reference currents; PhiGRL minimizes the
physics-informed loss defined in the original work; and
U-PINet minimizes the point-collocation EFIE residual
in~\eqref{eq:physics_loss}. The Adam optimizer is used
for all models with an initial learning rate
of $5\times10^{-4}$, a weight-decay coefficient
of $1\times10^{-5}$, a mini-batch size of $32$, and a
total of $300$ training epochs. Identical optimizer
settings are applied across all models, so that the
comparison reflects differences in architectural design
and training objective rather than in the training
budget. 

\section{Experimental Results and Discussion}
\label{section:E}

This section reports the surface-current and RCS results
obtained by U-PINet and compares them with the two
physics-informed baselines under the experimental setup
of Section~\ref{section:D}.

\subsection{Accuracy Validation on Canonical and Complex Targets}
\label{subsec:VI-A}

We first quantify the overall accuracy of all three methods
on the seven PEC targets. Table~\ref{tab:overall_accuracy}
reports the current-level errors
($\mathrm{NMSE}_{J}$, $\mathrm{MAE}_{J}$) and the
RCS-level errors ($\mathrm{RMSE}_{\mathrm{RCS}}$,
$R^{2}_{\mathrm{RCS}}$) with respect to the MLFMA
reference. Unless otherwise indicated, the entries in
Table~\ref{tab:overall_accuracy} are evaluated at the
operating frequency of $1$~GHz and averaged over the HH
and VV co-polarized configurations and over the
corresponding test incident and observation angles, so
that each table entry summarizes the target-level
behavior of the method under a matched excitation
setting.

\begin{figure}[t]
\centering
\includegraphics[width=9cm]{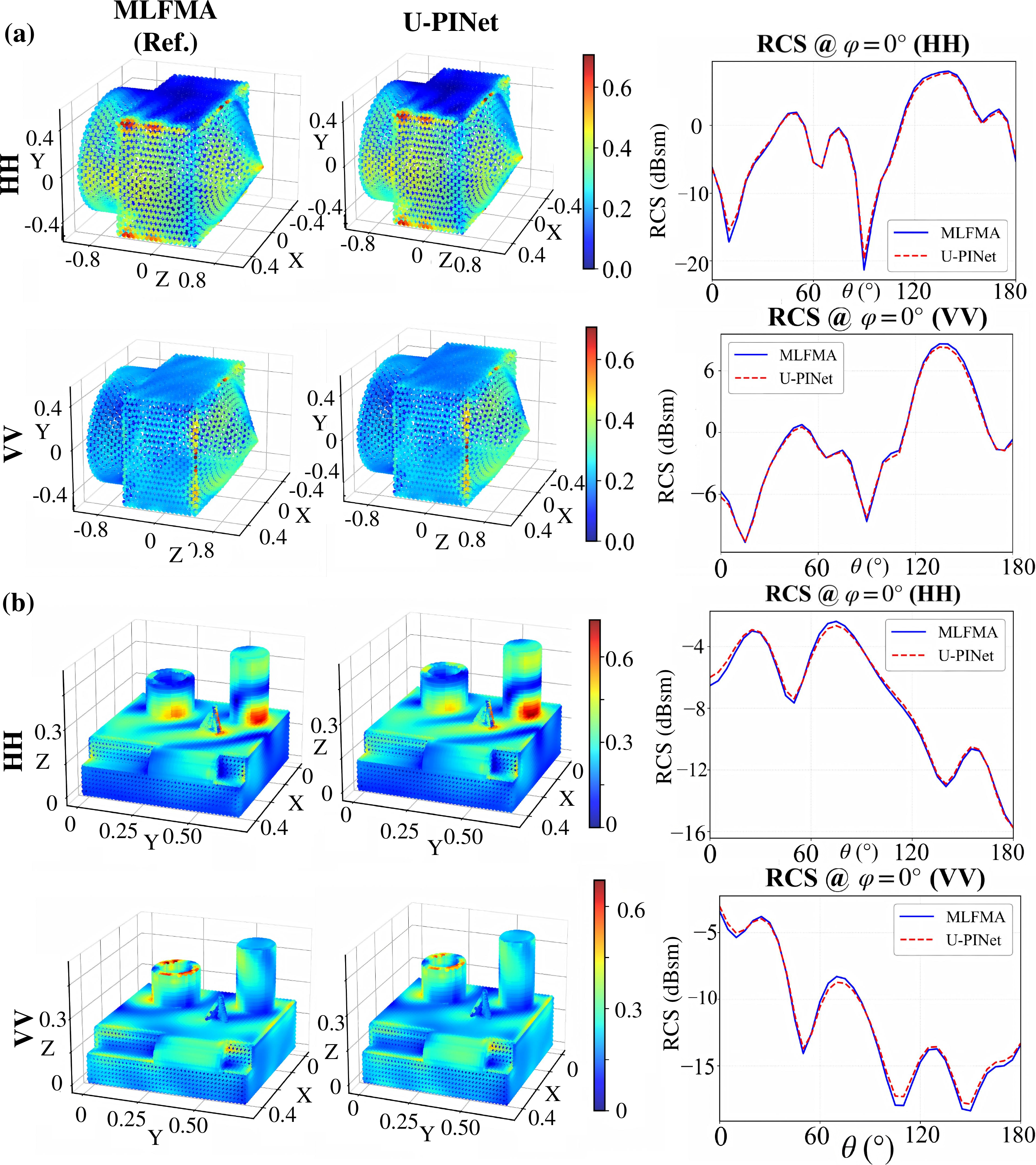}
\caption{Surface-current reconstruction and bistatic RCS prediction
for (a) cube and (b) SLICY. Within each subfigure, the upper and
lower rows correspond to HH and VV polarizations, as indicated by
the row labels. The MLFMA reference and the U-PINet prediction are
shown side by side, and the RCS curves are computed from the
corresponding reconstructed currents at $\varphi=0^\circ$.}
\label{fig: exfig1}
\end{figure}

\begin{figure*}[!t]
\centering
\includegraphics[width=18cm]{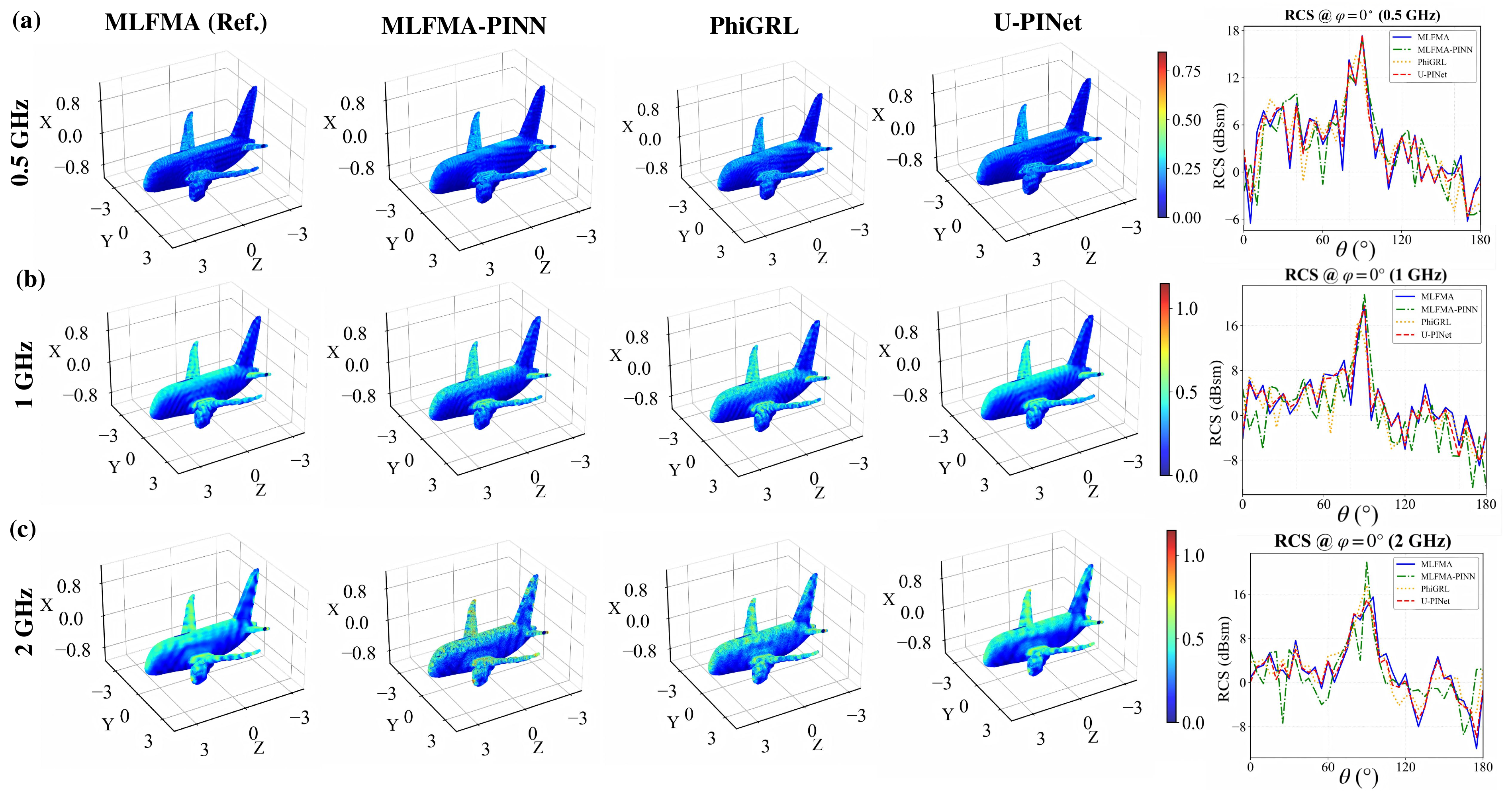}
\caption{Multi-frequency surface-current and RCS comparison on the
airplane target at $0.5$, $1$, and $2$~GHz, as indicated by the row labels.
The first column shows the MLFMA reference, followed by the
predictions of MLFMA-PINN, PhiGRL, and U-PINet; the last column
compares the corresponding bistatic RCS curves at $\varphi=0^\circ$.}
\label{fig: exfig2}
\end{figure*}

Across all seven targets, U-PINet attains the lowest
current-level and RCS-level errors and the highest
$R^{2}_{\mathrm{RCS}}$. Averaged over targets, U-PINet
reduces $\mathrm{NMSE}_{J}$ by roughly a factor of three
relative to MLFMA-PINN
 and by
about $46\%$ relative to PhiGRL, while
the RCS RMSE drops from $1.60$~dB and $1.41$~dB,
respectively, to $1.28$~dB. The trend is consistent across
both metric families, indicating that the current-level
improvement carries over to the far-field response.

The advantage of U-PINet becomes more pronounced as the
target complexity increases. For the simplest cube target,
all three methods stay within $1$~dB of RCS RMSE and above
$0.95$ of $R^{2}_{\mathrm{RCS}}$, reflecting the relatively
short-range nature of the dominant scattering. As the
geometry becomes more involved, the baseline errors grow
considerably faster than those of U-PINet. On the ship,
which is the most demanding case, MLFMA-PINN and PhiGRL
yield $\mathrm{NMSE}_{J}$ values of $1.62\!\times\!10^{-2}$
and $1.05\!\times\!10^{-2}$ and $R^{2}_{\mathrm{RCS}}$
values of $0.712$ and $0.814$, while U-PINet achieves
$5.65\!\times\!10^{-3}$ and $0.865$, respectively. This is
consistent with the physical intuition that, for
electrically larger and geometrically complex targets, both
short-range impedance coupling and long-range radiation
coupling contribute significantly to the induced current. This implies that a model that addresses both contributions explicitly
benefits the most.

\begin{figure*}[t]
\centering
\includegraphics[width=18cm]{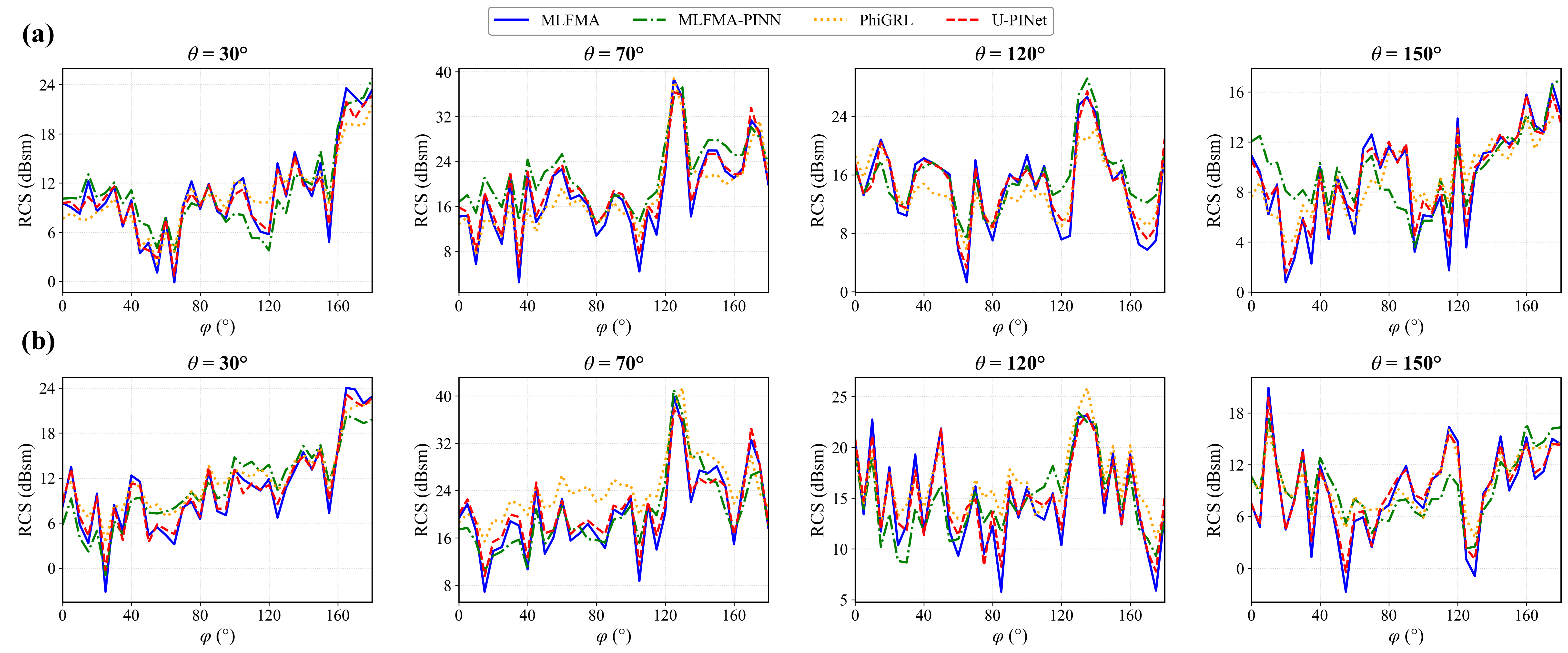}
\caption{Bistatic RCS comparison on the ship target under
HH (upper row) and VV (lower row) polarizations. Each
column shows the RCS variation with~$\varphi$ at a fixed
observation angle~$\theta$. The MLFMA result is used as the
reference, and the learning-based methods are compared
against it over multiple angular cuts.}
\label{fig: exfig3}
\end{figure*}

\begin{table*}[!t]
\centering
\caption{Configuration-Level Generalization Comparison
Under Held-Out Frequency, Polarization, and
Observation-Angle Settings.}
\label{tab:config_generalization}
\renewcommand{\arraystretch}{1.10}
\setlength{\tabcolsep}{3.5pt}
\footnotesize
\begin{adjustbox}{max width=\textwidth}
\begin{tabular}{llp{3.2cm}cccccc}
\toprule
\multirow{2}{*}{Test Type}
& \multirow{2}{*}{Target}
& \multirow{2}{*}{Held-Out Test Setting}
& \multicolumn{2}{c}{MLFMA-PINN}
& \multicolumn{2}{c}{PhiGRL}
& \multicolumn{2}{c}{U-PINet (Proposed)} \\
\cmidrule(lr){4-5}
\cmidrule(lr){6-7}
\cmidrule(lr){8-9}
& & 
& $\mathrm{NMSE}_{J}\downarrow$
& $\mathrm{RMSE}_{\mathrm{RCS}}\downarrow$
& $\mathrm{NMSE}_{J}\downarrow$
& $\mathrm{RMSE}_{\mathrm{RCS}}\downarrow$
& $\mathrm{NMSE}_{J}\downarrow$
& $\mathrm{RMSE}_{\mathrm{RCS}}\downarrow$ \\
\midrule

\multirow{3}{*}{Frequency}
& Airplane & 0.5 GHz
& $9.43 \times 10^{-3}$ & 1.97 & $6.18 \times 10^{-3}$ & 1.63 & $3.76 \times 10^{-3}$ & 1.41 \\
& Airplane & 1.0 GHz
& $1.28 \times 10^{-2}$ & 2.64 & $8.42 \times 10^{-3}$ & 2.31 & $5.00 \times 10^{-3}$ & 2.15 \\
& Airplane & 2.0 GHz
& $1.67 \times 10^{-2}$ & 3.46 & $1.13 \times 10^{-2}$ & 2.91 & $6.48 \times 10^{-3}$ & 2.63 \\

\midrule
\multirow{4}{*}{Polarization}
& SLICY & HH polarization
& $8.48 \times 10^{-3}$ & 1.07 & $5.81 \times 10^{-3}$ & 0.99 & $3.18 \times 10^{-3}$ & 0.94 \\
& SLICY & VV polarization
& $8.36 \times 10^{-3}$ & 1.01 & $5.71 \times 10^{-3}$ & 0.93 & $3.12 \times 10^{-3}$ & 0.90 \\
& Ship & HH polarization
& $1.65 \times 10^{-2}$ & 3.68 & $1.08 \times 10^{-2}$ & 3.27 & $5.73 \times 10^{-3}$ & 2.98 \\
& Ship & VV polarization
& $1.59 \times 10^{-2}$ & 3.56 & $1.02 \times 10^{-2}$ & 3.15 & $5.57 \times 10^{-3}$ & 2.92 \\

\midrule
\multirow{4}{*}{Observation Angle}
& Ship & $\theta \in [0^\circ,45^\circ]$
& $1.57 \times 10^{-2}$ & 3.51 & $9.80 \times 10^{-3}$ & 3.12 & $5.32 \times 10^{-3}$ & 2.84 \\
& Ship & $\theta \in [45^\circ,90^\circ]$
& $1.63 \times 10^{-2}$ & 3.68 & $1.07 \times 10^{-2}$ & 3.24 & $5.77 \times 10^{-3}$ & 2.98 \\
& Ship & $\theta \in [90^\circ,135^\circ]$
& $1.69 \times 10^{-2}$ & 3.74 & $1.11 \times 10^{-2}$ & 3.33 & $5.84 \times 10^{-3}$ & 3.07 \\
& Ship & $\theta \in [135^\circ,180^\circ]$
& $1.59 \times 10^{-2}$ & 3.55 & $1.04 \times 10^{-2}$ & 3.15 & $5.67 \times 10^{-3}$ & 2.91 \\
\bottomrule

\end{tabular}
\end{adjustbox}

\vspace{0.5mm}
\footnotesize{
Note: Each block isolates one held-out factor while
fixing the others. The Frequency block reports
errors averaged over HH and VV polarizations at each
listed frequency; the Polarization block is
evaluated at $1$~GHz; the Observation Angle
block is evaluated at $1$~GHz and averaged over HH and
VV polarizations. All errors are computed with respect
to the MLFMA reference under matched excitation and
observation settings.
}
\end{table*}

Figure~\ref{fig: exfig1} shows representative visual
comparisons on the cube and SLICY targets under HH and VV
polarizations. For the cube, U-PINet recovers the
concentration of surface current along the edges and
corners, where geometric discontinuities induce strong
local variations. For SLICY, which combines flat plates,
cylinders, and localized protrusions, the reconstructed
current preserves the dominant hotspots and their spatial
distribution on each substructure. The RCS curves derived
from the reconstructed currents track the MLFMA reference
under both polarizations across the swept $\theta$ range,
indicating that current-level fidelity translates into
consistent far-field behavior. The differences between HH
and VV cases are minor, suggesting that the model is not
overfitted to a particular polarization.

Figure~\ref{fig: exfig2} extends the comparison to the
airplane target at three frequencies. The airplane geometry
involves multi-scale features (wings, fuselage, engines,
tail), and as the frequency increases from $0.5$ to
$2$~GHz, the electrical size of the target grows and the
angular RCS pattern becomes more oscillatory. All three
methods see their RCS RMSE increase with frequency, but
U-PINet shows the most gradual degradation:
All three methods see their RCS RMSE increase with frequency,
but U-PINet shows the most gradual degradation: averaged
across the three test frequencies, its $R^2_{\mathrm{RCS}}$
remains at $0.915$, whereas MLFMA-PINN and PhiGRL drop to
$0.804$ and $0.871$, respectively. Visually, U-PINet reproduces the dominant
peaks and nulls of the bistatic RCS curve at all three
frequencies, while the baselines exhibit visible amplitude
mismatches and small angular shifts of the peaks. This
observation is consistent with the fact that the
hierarchical far-field branch is designed to capture
multi-scale coupling, which becomes more important as the
target becomes electrically larger.

Figure~\ref{fig: exfig3} further examines the ship target,
which is the longest and most aspect-dependent case in the
study. The RCS is plotted against~$\varphi$ at four fixed
observation angles ($\theta = 30^{\circ}$, $70^{\circ}$,
$120^{\circ}$, $150^{\circ}$) for both HH and VV
polarizations. U-PINet follows the MLFMA reference closely
in all eight cuts, including around the deep nulls between
$\varphi = 40^{\circ}$ and $80^{\circ}$ and the sharp peaks
near $\varphi = 120^{\circ}$--$140^{\circ}$, where the
elongated hull and superstructure produce strong
directional scattering. In contrast, MLFMA-PINN tends to
overshoot in the high-RCS regions and PhiGRL underestimates
several of the deep nulls, leading to visible amplitude
bias at multiple angular cuts. These observations match the
quantitative gap reported in
Table~\ref{tab:overall_accuracy}, where U-PINet improves
$\mathrm{RMSE}_{\mathrm{RCS}}$ on the ship by $0.67$~dB
over MLFMA-PINN and by $0.26$~dB over PhiGRL.

To complement the target-wise averages in
Table~\ref{tab:overall_accuracy},
Table~\ref{tab:config_generalization} reports a configuration-level
comparison by varying frequency, polarization, and observation-angle
range separately while keeping the remaining factors fixed. For the
airplane target, the errors of all methods increase as the frequency
rises from $0.5$ to $2$~GHz, but U-PINet shows the mildest
degradation. This frequency-induced growth is consistent with the
fact that an increasing electrical size shortens the wavelength
relative to the target, so that the induced current and the bistatic
RCS pattern become more oscillatory and contain finer-scale spatial
features that are intrinsically harder to reconstruct under a fixed network capacity. In the polarization tests on SLICY and
the ship, the HH--VV performance gap remains limited, while U-PINet
consistently achieves the lowest current and RCS errors, indicating
that its accuracy advantage is not tied to a specific co-polarized
excitation. For the ship angular-sector test, all methods exhibit
the largest errors in the $\theta\in[90^{\circ},135^{\circ}]$ sector,
where the elongated hull and superstructure produce a more
directional and oscillatory scattering response; even in this most
demanding sector, U-PINet retains the lowest RCS error among the
three methods. Overall, the advantage of U-PINet is preserved across
frequency, polarization, and angular sectors, with the gap over the
baselines widening under electrically larger and directionally
complex conditions.

The visual comparisons in Figs.~\ref{fig: exfig1}%
--\ref{fig: exfig3} reinforce the quantitative trends
reported in
Tables~\ref{tab:overall_accuracy}--\ref{tab:config_generalization}:
U-PINet produces consistent improvements at both the
current and the RCS levels, and the margin over the
baselines widens with target complexity, operating
frequency, and the multi-scale character of the
scattering response.

The accuracy reported above is measured at the RCS
level, which is the metric relevant to most
repeated-query microwave scattering tasks. A
complementary, equation-level view of the predicted
current is examined in
Section~\ref{subsec:solver_validation}.

\subsection{Ablation Study}

To assess the individual contributions of the near-field
graph encoder and the hierarchical far-field fusion module,
we conduct a controlled ablation study by selectively
replacing each branch with a simpler counterpart and
retraining the model under the same EFIE residual objective
of~\eqref{eq:physics_loss}. The study is reported on the
ship target, which is the most demanding case in
Table~\ref{tab:overall_accuracy} and therefore produces the
largest and most readable gaps between variants. The same
ablation protocol has been performed on the airplane and
SLICY targets, and the relative ordering of variants is
consistent across these targets. Three variants of
U-PINet are considered:

\begin{itemize}
\item \textbf{w/o near-field graph}: the
graph-based near-field encoder of
Section~\ref{subsec:nf_encoder} is replaced by a per-point
 MLP that processes each surface
point independently. This setting removes the explicit
modeling of pairwise short-range interactions on the surface
graph while keeping the per-point input features unchanged.

\item \textbf{w/o far-field hierarchy}: the hierarchical
fusion module of
Section~\ref{sec:far_field_module} is replaced by a
global fully connected layer acting directly on the
finest-level features at $k = L$, with no pooling, no
cross-scale exchange, and no coarse-to-fine fusion. This
setting removes the multi-scale structure used to capture
long-range coupling.

\item \textbf{w/o near--far modeling}: both branches are
replaced as above, yielding a baseline that combines a
per-point MLP encoder with a single global fully connected
mapping at the finest level.
\end{itemize}

The results are summarized in Table~\ref{tab:ablation}.
Both single-branch ablations degrade the accuracy at the
current and the RCS levels relative to the full U-PINet.
Removing the near-field graph (``w/o near-field graph'')
increases $\mathrm{NMSE}_{J}$ from $5.65\!\times\!10^{-3}$
to $7.95\!\times\!10^{-3}$ and reduces $R^{2}_{\mathrm{RCS}}$
from $0.865$ to $0.812$, indicating that the per-point MLP
alone cannot adequately represent current variations driven
by short-range coupling between neighboring mesh elements,
particularly near edges and corners of the ship hull.
Removing the hierarchical far-field branch (``w/o far-field
hierarchy'') produces a comparable current-level error but
a noticeably larger RCS-level degradation,
$\mathrm{RMSE}_{\mathrm{RCS}}$ rising from $2.95$ to
$3.76$~dB and $R^{2}_{\mathrm{RCS}}$ dropping to $0.785$.
This asymmetry is consistent with the role of the multi-scale
branch in modeling long-range radiation coupling: such
coupling has a relatively localized effect on the surface
current but a pronounced effect on the bistatic far-field
response of an electrically large and elongated target.
Removing both branches simultaneously (``w/o near--far
modeling'') yields the worst result across all four metrics,
with $\mathrm{NMSE}_{J}$ roughly twice that of the full
model and $R^{2}_{\mathrm{RCS}}$ falling below $0.73$. This
result indicates that the two design choices are
complementary rather than redundant, with the near-field
branch contributing primarily to current-level accuracy and
the far-field branch contributing primarily to far-field
scattering consistency.

\begin{table}[!t]
\centering
\caption{Ablation Study of the Near-Field and Far-Field Modules on the Ship Target}
\label{tab:ablation}
\renewcommand{\arraystretch}{1.10}
\setlength{\tabcolsep}{3.5pt}
\small
\begin{tabular}{lcccc}
\toprule
Model Variant 
& $\mathrm{NMSE}_{J}^{\downarrow}$ 
& $\mathrm{MAE}_{J}^{\downarrow}$ 
& $\mathrm{RMSE}_{\mathrm{RCS}}^{\downarrow}$ 
& $R^{2,\uparrow}_{\mathrm{RCS}}$ \\
\midrule
Full U-PINet 
& $5.65 \times 10^{-3}$ & 0.0671 & 2.95 & 0.865 \\

w/o near-field graph 
& $7.95 \times 10^{-3}$ & 0.0882 & 3.38 & 0.812 \\

w/o far-field hierarchy 
& $7.15 \times 10^{-3}$ & 0.0795 & 3.76 & 0.785 \\

w/o near--far modeling 
& $1.18 \times 10^{-2}$ & 0.1085 & 4.25 & 0.728 \\
\bottomrule
\end{tabular}

\vspace{0.5mm}
\footnotesize{
Note: ``w/o near-field graph'', ``w/o far-field hierarchy'', and
``w/o near--far modeling'' replace the graph encoder with a
per-point MLP, replace the hierarchical fusion module with a
finest-level fully connected layer, and apply both replacements,
respectively. $\downarrow$/$\uparrow$ denotes lower/higher is better,
and $\mathrm{RMSE}_{\mathrm{RCS}}$ is reported in dB.
}
\end{table}

\begin{table}[!t]
\centering
\caption{Performance of U-PINet-Assisted GMRES Initialization on Three Complex Targets.}
\label{tab:gmres_init}
\renewcommand{\arraystretch}{1.10}
\setlength{\tabcolsep}{3pt}
\footnotesize
\resizebox{\columnwidth}{!}{%
\begin{tabular}{llccccc}
\toprule
\textbf{Target}
& \textbf{Init.}
& $\eta_0\downarrow$
& \textbf{Iter.}$\downarrow$
& $t_{\mathrm{iter}}$~(s)$\downarrow$
& $t_{\mathrm{online}}$~(s)$\downarrow$
& \textbf{Red.}$\uparrow$ \\
\midrule
\multirow{4}{*}{SLICY}
& Zero        & 1.000 & 185 & 45.2   & 45.2   & --     \\
& MLFMA-PINN  & 0.450 & 110 & 26.8   & 28.5   & 40.5\% \\
& PhiGRL      & 0.320 & 82  & 20.0   & 21.8   & 55.7\% \\
& U-PINet     & 0.150 & 45  & 11.0   & 12.5   & 75.7\% \\
\midrule
\multirow{4}{*}{Airplane}
& Zero        & 1.000 & 420 & 655.0  & 655.0  & --     \\
& MLFMA-PINN  & 0.620 & 310 & 483.5  & 487.2  & 26.2\% \\
& PhiGRL      & 0.510 & 245 & 382.0  & 386.5  & 41.7\% \\
& U-PINet     & 0.280 & 130 & 202.8  & 207.5  & 69.0\% \\
\midrule
\multirow{4}{*}{Ship}
& Zero        & 1.000 & 540 & 2450.0 & 2450.0 & --     \\
& MLFMA-PINN  & 0.750 & 430 & 1950.0 & 1965.5 & 20.4\% \\
& PhiGRL      & 0.620 & 350 & 1585.0 & 1598.2 & 35.2\% \\
& U-PINet     & 0.350 & 210 & 950.0  & 958.5  & 61.1\% \\
\bottomrule
\end{tabular}%
}
\vspace{0.5mm}
\footnotesize{
Note: The GMRES convergence tolerance is $10^{-3}$.
$\eta_0$ is the normalized initial residual defined
in~\eqref{eq:initial_residual};
$t_{\mathrm{iter}}$ is the GMRES iteration time;
$t_{\mathrm{online}}$ also includes the network inference
time required to produce $\mathbf{I}_0$ via the projection
in~\eqref{eq:rwg_projection};
``Red.'' denotes the iteration-count reduction relative
to zero initialization.
$\downarrow$/$\uparrow$ indicate that lower/higher values
are better. All learning-based initializers are evaluated
under the same per-target GMRES protocol.
}
\end{table}

\subsection{Solver-Oriented Validation}
\label{subsec:solver_validation}

The accuracy reported in Section~\ref{subsec:VI-A} is measured
at the RCS level, which is determined by the radiation integral
in~\eqref{eq:Esca_operator} and therefore tolerates small,
spatially distributed errors in the predicted current that are
averaged out during far-field integration. A complementary
question is how the same predicted current behaves when measured
at the equation level, that is, as a candidate solution of the
discretized linear system $\mathbf{ZI}=\mathbf{V}$. This
distinction is relevant because certain downstream tasks, such
as near-field analysis, EMC evaluation involving coupled
sub-systems, and iterative inverse-problem pipelines, require
the surface current to satisfy the discretized EFIE to a
numerical tolerance that is much stricter than the one needed
for accurate bistatic RCS reconstruction.

To examine this regime, we evaluate U-PINet not as a replacement
of the classical solver but as a source of physics-consistent
initial guesses for the iterative EFIE solver. The experiment is
conducted per target, so the goal is not to demonstrate
cross-shape transfer, but to quantify how much iterative effort
can be saved on the same target by warm-starting GMRES with a
learned current.

Let $\mathbf{I}_0$ denote the initial coefficient vector supplied
to GMRES. For the zero-initialization case,
$\mathbf{I}_0 = \mathbf{0}$. For the learning-based
initializations, the predicted surface current
$\boldsymbol{J}_{\mathrm{pred}}(\boldsymbol{p}_i)$
at each mesh-element centroid is projected onto the RWG
basis $\{\boldsymbol{f}_n\}_{n=1}^{N}$ defined on the same
triangular mesh through a standard least-squares fit,
\begin{equation}
    \mathbf{I}_0
    =
    \arg\min_{\mathbf{I}\in\mathbb{C}^{N}}
    \sum_{i}
    \Big\|
        \boldsymbol{J}_{\mathrm{pred}}(\boldsymbol{p}_i)
        -
        \sum_{n=1}^{N} I_n\,\boldsymbol{f}_n(\boldsymbol{p}_i)
    \Big\|^{2},
    \label{eq:rwg_projection}
\end{equation}
yielding the coefficient vector $\mathbf{I}_0$ used to
warm-start the iterative solver. The normalized initial
residual is defined as
\begin{equation}
    \eta_0 =
    \frac{\left\|\mathbf{V}-\mathbf{Z}\mathbf{I}_0\right\|_2}
    {\left\|\mathbf{V}\right\|_2},
    \label{eq:initial_residual}
\end{equation}
where $\mathbf{Z}$ and $\mathbf{V}$ are the MoM impedance matrix
and the tested incident-field vector from
Section~\ref{subsec:mlfma}. The same MLFMA-accelerated GMRES
implementation is used for all initializations, with the
convergence criterion $\|\mathbf{V}-\mathbf{Z}\mathbf{I}\|_2 /
\|\mathbf{V}\|_2 < 10^{-3}$.

The results are reported in Table~\ref{tab:gmres_init}. The
learning-based initializations all start from a normalized
residual well below unity, indicating that the predicted
currents are physically consistent with the EFIE operator even
though they do not satisfy the discretized system at the
$10^{-3}$ tolerance. Among the three methods, U-PINet attains
the lowest $\eta_0$ on every target, which translates into the
smallest iteration count and the largest reduction in online
solution time. We note that the iteration-count reduction is
not simply proportional to $\log\eta_0 / \log(\mathrm{tol})$,
since GMRES convergence is governed by the spectral
distribution of $\mathbf{Z}$ within the Krylov subspace
generated by the initial residual rather than by the residual
magnitude alone. The observed reductions therefore reflect, in
addition to a smaller starting residual, the extent to which
the learned current aligns with the dominant Krylov directions
of the discretized EFIE operator.

These observations clarify the role of U-PINet in two distinct
operating modes. When the downstream task is bistatic RCS
evaluation, the standalone surrogate in
Section~\ref{subsec:VI-A} is sufficient, and no iterative
correction is required. When the downstream task requires the
surface current to satisfy the discretized EFIE to a numerical
tolerance, the same predicted current can be reused as an
initial guess for a classical MLFMA-accelerated GMRES solver,
reducing the iteration count without modifying the underlying
solver. The two modes are complementary, and the choice between
them is governed by the accuracy required by the downstream
task rather than by a limitation of the surrogate itself.

\begin{table}[!t]
\centering
\caption{Runtime Comparison for Multi-Angle RCS Evaluation.}
\label{tab:runtime}
\renewcommand{\arraystretch}{1.12}
\setlength{\tabcolsep}{3.4pt}
\scriptsize
\resizebox{\columnwidth}{!}{%
\begin{tabular}{lrccc}
\hline
\textbf{Target} &
\textbf{Mesh Elem.} &
\textbf{MLFMA} &
\textbf{U-PINet Train} &
\textbf{U-PINet Infer.} \\
& &
\textbf{($360$ angles, h)} &
\textbf{(h)} &
\textbf{($360$ angles, s)} \\
\hline
SLICY    & $3{,}080$  & $4.52$   & $1.5$  & $3.5$  \\
Assembly & $4{,}830$  & $7.50$   & $2.0$  & $4.8$  \\
Sphere   & $8{,}050$  & $13.50$  & $2.8$  & $7.2$  \\
Cube     & $12{,}380$ & $22.00$  & $3.5$  & $10.5$ \\
Cone     & $13{,}090$ & $24.50$  & $3.8$  & $12.8$ \\
Airplane & $39{,}100$ & $65.50$  & $8.5$  & $25.5$ \\
Ship     & $78{,}460$ & $245.00$ & $26.0$ & $52.4$ \\
\hline
\end{tabular}%
}

\vspace{0.5mm}
\footnotesize{
Note: All timings refer to $360$ incident-angle
configurations on the same target. The U-PINet training
time is the per-target offline cost, incurred only once.
}
\end{table}

\subsection{Computational Efficiency}
\label{subsec:efficiency}

We finally compare the wall-clock cost of MLFMA and
U-PINet in the multi-angle RCS evaluation scenario, which
is one of the practical use cases motivating the proposed
surrogate. For MLFMA, the reported time corresponds to the
cumulative iterative solution time required to compute the
RCS at $360$ incident-angle configurations on the same
target. For U-PINet, the offline training cost and the
online inference cost are reported separately, with the
inference time measured as a single forward pass per
incident angle, summed over the same $360$ configurations.

Table~\ref{tab:runtime} summarizes the comparison. The
online inference cost of U-PINet is several orders of
magnitude lower than the cumulative MLFMA solution time on
every target, and the gap widens with the mesh size
because the per-angle MLFMA cost grows with the problem
size whereas the per-angle inference cost grows much more
slowly. The offline training cost ranges from a few hours
to about $26$~h depending on the target, and is incurred
only once. Because the training cost is amortized over all
subsequent queries on the same target, the total cost of
U-PINet remains substantially below that of MLFMA already
after a relatively small number of incidence
configurations. The proposed surrogate is therefore well
suited to repeated-query scattering analysis on a fixed
geometry, in which the same target is evaluated under many
incidence directions, polarizations, or frequencies.

\section{Conclusions}
\label{section:conclusion}

This paper introduced U-PINet, a physics-informed
hierarchical neural network for 3D microwave scattering modeling
of PEC targets. Motivated by the near--far separation of
length scales in the underlying MLFMA discretization,
U-PINet combines a near-field graph encoder, whose
message passing on the surface mesh is parameterized by
learnable univariate basis functions, with a hierarchical
far-field fusion module organized on an octree partition.
A discretized EFIE residual loss is used as the sole
training objective, allowing the proposed network to learn the
induced surface current without reference current labels
from a numerical solver.

The proposed framework was evaluated on seven canonical and
geometrically complex 3D PEC targets at microwave
frequencies. Compared with two representative
physics-informed CEM baselines, it was showcased that U-PINet attains
consistently lower current-level and bistatic RCS errors
across all targets, with the margin widening on
electrically larger and multi-scale geometries, such as the
airplane and the ship. The conducted ablation study indicated that
the two branches play complementary roles: the near-field
branch contributes primarily to current-level fidelity,
whereas the hierarchical far-field branch contributes
primarily to the consistency of the far-field response.
Beyond its use as a standalone surrogate, it was demonstrated that U-PINet also acts as an informed initializer for the classical
MLFMA-accelerated GMRES solver, reducing the iteration
count and the online solution time on every target tested.
Together, these results suggest that an architecture aligned
with the physical structure of the integral-equation
discretization, rather than a generic neural model, can
offer practical benefits for repeated-query microwave
scattering analysis on a fixed geometry.

Future work will focus on extending U-PINet to broadband RCS analysis through frequency-conditioned network formulations, and on incorporating dielectric and impedance boundary conditions to broaden its applicability beyond PEC targets. Application of the framework to downstream radar signal processing tasks, including synthetic aperture radar (SAR) imaging and automatic target recognition, where rapid generation of high-fidelity scattering signatures is critical, will also be pursued.

\bibliographystyle{IEEEtran}
\bibliography{IEEEabrv,refs}

\end{document}